\documentclass[preprint,12pt,authoryear]{elsarticle}

\usepackage{amssymb}
\usepackage{enumitem} % 放在导言区
\usepackage{amsmath}
\usepackage{bbm}
\usepackage{bm}
\usepackage{booktabs}
\usepackage{graphicx}
\usepackage{subcaption}
\graphicspath{{figures/}}

\journal{Neural Networks}

\begin{document}

\begin{frontmatter}

%% Title, authors and addresses

%% use the tnoteref command within \title for footnotes;
%% use the tnotetext command for theassociated footnote;
%% use the fnref command within \author or \affiliation for footnotes;
%% use the fntext command for theassociated footnote;
%% use the corref command within \author for corresponding author footnotes;
%% use the cortext command for theassociated footnote;
%% use the ead command for the email address,
%% and the form \ead[url] for the home page:
%% \title{Title\tnoteref{label1}}
%% \tnotetext[label1]{}
%% \author{Name\corref{cor1}\fnref{label2}}
%% \ead{email address}
%% \ead[url]{home page}
%% \fntext[label2]{}
%% \cortext[cor1]{}
%% \affiliation{organization={},
%%            addressline={}, 
%%            city={},
%%            postcode={}, 
%%            state={},
%%            country={}}
%% \fntext[label3]{}

\title{Degraded but Not Entirely Ineffective: PE-Based Deformable Graph Neural Networks} %% Article title

%% use optional labels to link authors explicitly to addresses:
%% \author[label1,label2]{}
%% \affiliation[label1]{organization={},
%%             addressline={},
%%             city={},
%%             postcode={},
%%             state={},
%%             country={}}
%%
%% \affiliation[label2]{organization={},
%%             addressline={},
%%             city={},
%%             postcode={},
%%             state={},
%%             country={}}

%\author{Wu Jin Hua} %% Author name
% 作者1：非通讯作者
% 作者1：非通讯作者
% 作者1：非通讯作者，无需写邮箱
\author[aff1]{Jinhua Wu}

% 作者2：通讯作者，\corref 和 \ead 都嵌在作者大括号内，全局只写这一次
\author [aff1]{Xinliang Zhang\corref {cor1}}
\ead {zxldq@hpu.edu.cn}  % 换成真实邮箱
% 单位：结构化写法，内部绝对不要出现 \ead 命令
\affiliation[aff1]{
    organization={Henan Polytechnic University},
    addressline={},
    city={Jiaozuo},
    postcode={454000},
    state={Henan},
    country={China}
}
% 通讯作者脚注：只写文字说明，不要加 \ead，也不要手动写邮箱
\cortext[cor1]{Corresponding author}
%%\cortext[cor1]{Corresponding author: Xinliang Zhang}

%% Abstract
\begin{abstract}
%% Text of abstract
Many real-world scenarios can be represented using graph-structured data. However, traditional GNNs that transmit messages based on first-order neighbors have long faced several fundamental contradictions: increasing depth leads to over-smoothing, long-range dependencies cause over-compression, fixed neighborhoods restrict the receptive field, and on heterophilous graphs, topological neighbors become a source of noise. Although many works have addressed these issues individually, few mechanisms can simultaneously alleviate all of these challenges. To address the aforementioned problems, we propose a Position Encoding-Based Deformable Spatial Aggregation Module (PEBDSAM) that solves them all in one step. Specifically, we utilize a deformable mechanism in the position space to identify relevant nodes to supplement the original first-order neighbor information of GNNs, allowing traditional GNNs to adapt to heterophilous scenarios. Through diagnostic experiments, we obtained several major findings: current offsets fail to have any effect; subsequently, we analyzed the causes of offset failure and why model performance still improves even after offset failure, pointing out future research directions. Based on these diagnostic experiments, we streamlined the original PEBDSAM, resulting in a simplified version, which we call the Position Encoding-Based Spatial Aggregation Module (PEBSAM). In addition, we propose a PEBSAM-Speed to adapt to large datasets. Finally, we designed the module to be plug-and-play and applied it to GCN, GAT, GIN, and GraphSAGE, achieving desirable results on three homophilous datasets and six heterophilous graph datasets.
\end{abstract}

%%Graphical abstract
%\begin{graphicalabstract}
%\includegraphics{grabs}
%\end{graphicalabstract}

%%Research highlights
%\begin{highlights}
%\item Research highlight 1
%\item Research highlight 2
%\end{highlights}

%% Keywords
\begin{keyword}
Graph neural networks \sep Position encoding  \sep  Deformable  \sep  Diagnostic experiments  \sep  Heterophilous graph

%% PACS codes here, in the form: \PACS code \sep code

%% MSC codes here, in the form: \MSC code \sep code
%% or \MSC[2008] code \sep code (2000 is the default)

\end{keyword}

\end{frontmatter}

%% Add \usepackage{lineno} before \begin{document} and uncomment 
%% following line to enable line numbers
%% \linenumbers

%% main text
%%

%% Use \section commands to start a section
\section{Introduction}
\label{sec:intro}
%% Labels are used to cross-reference an item using \ref command.

Many real-world scenarios can be represented using graph-structured data, such as transportation networks, molecular structures, and social systems. Therefore, designing methods for processing graph-structured data is particularly important.

Early on, inspired by the tremendous success of convolutional neural networks in NLP and CV, \citet{Bruna2013Spectral} introduced convolution to graph-structured data for the first time, proposing spectral-based graph convolutional networks. At that time, defining graph convolution directly in the spatial domain was extremely difficult and challenging; therefore, they processed the graph signal in the frequency domain by computing all eigenvalues of the graph Laplacian matrix L, thereby defining the convolution operation in the spectral domain. To alleviate the significant computational cost associated with the expensive eigen decomposition required by spectral GCNs, \citet{defferrard2016convolutional} proposed ChebNet, using Chebyshev polynomials to approximate the spectral convolution kernel, thus avoiding eigen decomposition and making the model complexity linear with respect to the number of edges in the graph. \citet{kipf2017semi} went further by performing a first-order approximation of ChebNet, greatly simplifying the model. From a spatial perspective, it only aggregates first-order neighbors. However, the aggregation weights are fixed and not learnable. \citet{hamilton2017inductive} introduced GraphSAGE, which greatly reduced computational costs by sampling first-order neighbors, making it very suitable for large-scale graph datasets. Additionally, GraphSAGE introduced learnable aggregation methods, such as LSTM aggregation. With the advent of the Transformer, the status of LSTM has been significantly impacted. \citet{velickovic2018graph}, inspired by the attention mechanism in the Transformer, proposed GAT based on self-attention mechanisms, offering a more precise evaluation of neighbor importance. These methods build upon and improve previous work, not only featuring elegant and concise formulations, but also being widely applied and achieving remarkable success in various fields.

However, they suffer from several inherent problems:
\begin{enumerate}
    \item \textbf{Limited Receptive Field}: A single layer only has a first-order neighborhood receptive field. To expand it, stacking multiple layers is necessary, which not only increases the number of model parameters but also risks oversmoothing.
    \item \textbf{Over-squashing}: Nodes that are topologically distant in the graph require many GNN layers to obtain information from each other, and  step-by-step propagation across multiple edges leads to significant information loss.
    \item \textbf{Limited Generalizability}: Traditional GNNs are based on the homophily assumption, meaning that connected nodes tend to have similar features and labels. This makes them excel at processing homophilous graphs but often yields suboptimal performance on heterophilous graphs.
    \end{enumerate}
    
In computer vision (CV), \citet{Dai_2017_ICCV} proposed deformable convolutions. Traditional convolutional kernels have a fixed and regular shape, sampling at predetermined locations, and possessing a fixed receptive field size. This is analogous to current traditional GNNs : since the edges in static graph data are predefined and immutable, the single-layer receptive field of GNNs processing such data is also fixed—i.e., only first-order neighbors. Deformable convolution breaks the limitation of fixed regular grids, allowing the model to autonomously decide where to "look." To impart GNNs with the flexibility of deformable convolution, the model architecture must enable the adaptive adjustment of edges. This not only grants the model an adaptive receptive field but also allows for direct connections between originally distant nodes in the graph topology when the model adjusts the edges, thus avoiding over-compression. Adaptive adjustment and reconstruction of the edge structure also break the homophily assumption that if an edge exists between two nodes, they tend to have similar features and class labels. This makes the model applicable to heterophilic graphs, making it more generalizable. Thus, the aforementioned problems are solved in one stroke. As our work is still in the early exploratory stage, the designed modules are not yet perfected and could be improved further, as mentioned in Section~\ref{sec:dia}. At the same time, we hope that more people will join this community and explore together.

This paper proposes PEBDSAM. Specifically, traditional GNNs aggregate first-order neighbors through structural relationships; this information is referred to as structural information. Our PEBDSAM, on the other hand, provides spatial information. Each node within the position encoding space seeks new spatially relevant nodes through dynamic offsets and subsequently aggregates the features of these new nodes, which we refer to as spatial features. Finally, the model adaptively fuses structural and spatial features. The new nodes identified in the spatial features may be several hops away in the graph topology, allowing us to traverse long-range distances in the graph within a single aggregation, significantly alleviating oversquashing. However, these new nodes also provide a larger and more comprehensive receptive field, overcoming the limitation of first-order neighbors. To demonstrate the universality of our proposed model, we conducted experiments on both homophilic and heterophilic graphs, and our model exhibited superior performance. However, during subsequent model diagnostics, we discovered that the offsets in the deformable mechanism approached zero as the model converged, indicating the failure of the deformable mechanism. We provide an in-depth analysis of this phenomenon, revealing its underlying causes, as well as the current limitations of our work and suggestions for future research. Furthermore, we analyze why our model’s performance can improve even after the offsets fail. Finally, we report several additional findings. After conducting model diagnostics and based on these findings, we simplified our proposed PEBDSAM and introduced a streamlined model (PEBSAM). To enable our model to run on large-scale graphs, we further proposed PEBSAM-Speed. In summary, our main contributions are as follows:
\begin{enumerate}
    \item To address the poor performance of traditional GNNs on heterophilous graphs, we propose a novel architecture that combines structural and spatial information, thereby making standard GNNs applicable to heterophilous scenarios.
    \item In the field of graph neural networks, the role of position encoding is currently underestimated; therefore, we demonstrate a novel use case.
    \item To address the issues of fixed receptive field size and over-squashing of distant nodes in traditional GNNs, we propose PEBDSAM, which provides spatial information to simultaneously enlarge the receptive field and alleviate over-squashing.
    \item Our framework achieves promising performance on three homophilous graphs and six heterophilous graphs. Particularly for heterophilous graphs, performance improvements are pronounced, highlighting the strengths of our spatial information.
\end{enumerate}
We first review related work in Section~\ref{sec:related}, and then elaborate on our proposed method in Section~\ref{sec:method}. The diagnostic experiments and analyses are presented in Section~\ref{sec:dia}. A simplified model is presented in Section~\ref{sec:sim}. Experiments and ablation studies are reported in Section~\ref{sec:exp}, and conclusions are drawn in Section~\ref{sec:concl}.
\section{Related Work}
\label{sec:related} % 给相关工作打标签
\subsection{Deformable Convolution}

Convolutional neural networks have been widely applied and have achieved remarkable success in various fields. However, traditional convolutional kernels constrained by fixed receptive fields are limited in modeling objects with diverse deformations. To address this limitation, \citet{Dai_2017_ICCV,Zhu_2019_CVPR} proposed deformable convolution, which learns offsets to break the constraint of the fixed receptive field, enabling adaptive receptive fields. This technique was subsequently applied to feature alignment and numerous domains. Inspired by this work, it was incorporated into attention mechanisms, giving rise to deformable DETR\citep{zhu2021deformable}. The present work is devoted to introducing the deformable mechanism into GNNs to address the inherent flaws of conventional GNNs, such as fixed receptive fields.

\subsection{Existing Deformable Mechanisms in GNNs}

\citet{park2022deformable} proposed Deformable Graph Convolution Networks(DGCN), for which the core formula is as follows:
\begin{equation}
y_v = \sum_{u \in \tilde{N}(v)} g_{\text{deform}} \left( \mathbf{r}_{u,v}, \Delta_k \left( \mathbf{e}_v \right) \right) \mathbf{h}_u,
\label{eq:deform}
\end{equation}
From Eq.~\eqref{eq:deform}, it can be seen that the idea of deformable convolution is introduced into the aggregation approach. This novel aggregation method was shown in the original paper to outperform GAT, which is commendable. However, this deformable formula does not actually expand the receptive field, because $\mathbf{h}_u$ remains unchanged and still aggregates only first-order neighbors, with only the aggregation weights modified. Essentially, they expand the receptive field by constructing multiple latent neighbor graphs, as for any given node, its neighbors differ across various latent neighbor graphs. The final result is obtained by aggregating all neighbor graphs via attention.

\citet{park2025deformable} proposed a Deformable Graph Transformer (DGT). Their core innovation is to apply a NodeSort module that rearranges each node in the raw graph data according to multiple criteria, thereby generating several new and ordered node feature sequences. These are one-dimensional, akin to token sequences in NLP, and deformable graph attention (DGA) is then applied to these ordered sequences to achieve adaptive receptive fields. Both methods are highly generalizable and have achieved promising performance on both heterophilic and homophilic graphs, which is encouraging. However, our approach is distinct. Both aforementioned methods require preprocessing of the graph data—DGCN does so by constructing multiple latent neighbor graphs, whereas DGT rearranges node features using multiple criteria. In contrast, we introduce deformable convolution into the PE space using offsets to directly identify new and relevant nodes in the PE space that extend beyond the range of first-order neighbors, thereby expanding the receptive field and mitigating oversquashing.

\section{Proposed Method}
\label{sec:method}  % 给方法打标签

\subsection{Overall Architecture and Brief Analysis}

\begin{figure}[htb]
    \centering
    % 这里 your_image 不要带后缀名（如果有多个后缀，会按 eps/pdf/png 顺序匹配）
    \includegraphics[width=\columnwidth]{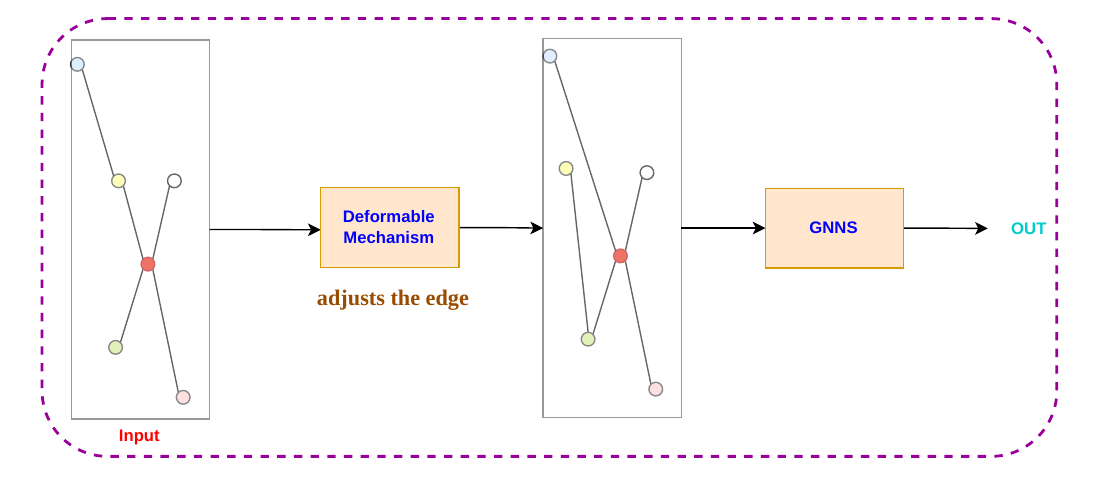}
    \caption{Overall architecture}
    \label{fig:over}
\end{figure}
The proposed model architecture is illustrated in Figure~\ref{fig:over}. Initially, our idea was to use a deformable mechanism to adjust the edges; however, in graph datasets, the number of edges significantly exceeds the number of nodes. Since operations on individual edges have high computational complexity, certain techniques must be employed. We also conducted preliminary experiments, which showed some performance improvement, but there remained a significant gap with the state-of-the-art (SOTA). \begin{figure}[htb]
    \centering
    % 这里 your_image 不要带后缀名（如果有多个后缀，会按 eps/pdf/png 顺序匹配）
    \includegraphics[width=\columnwidth]{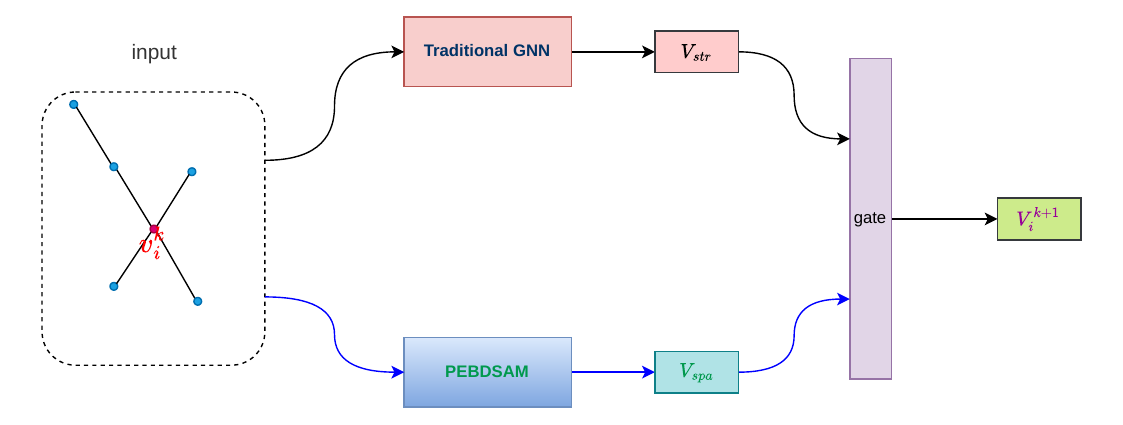}
    \caption{Illustration of the proposed Simplified Model}
    \label{fig:simple model}
\end{figure}
Therefore, we propose a simplified version (see Figure~\ref{fig:simple model}). PEBDSAM provides a larger receptive field; however, this may introduce unnecessary noise. The processing strategy is as follows: structural information and spatial information can be filtered through a gating mechanism to reduce noise in structural and spatial information, while within spatial information, an attention mechanism is used to diminish the influence of noisy nodes and enhance the role of relevant nodes. For structural information, traditional models, such as GAT, are also available to reduce the impact of noisy nodes. Therefore, this is theoretically feasible. In the following, we introduce the simplified model proposed in this paper.

\subsection{Model Introduction}

\begin{figure}[htb]
    \centering
    % 这里 your_image 不要带后缀名（如果有多个后缀，会按 eps/pdf/png 顺序匹配）Illustration of the proposed PEBDSAM
    \includegraphics[width=\columnwidth]{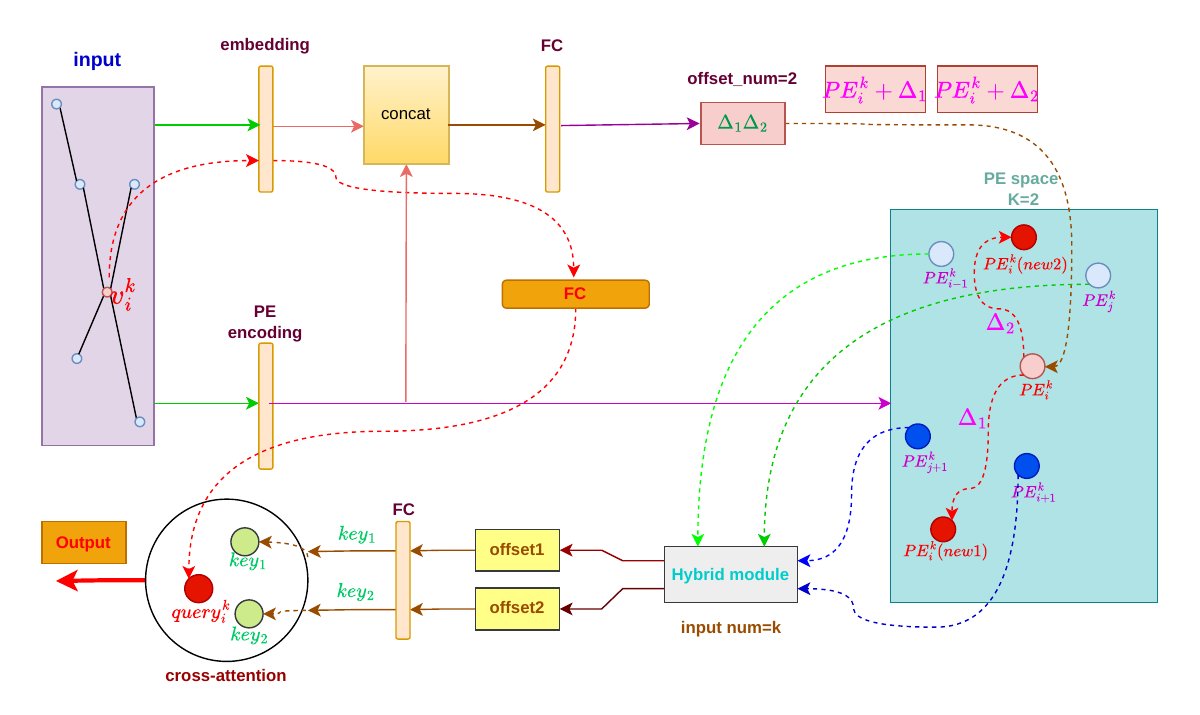}
    \caption{Illustration of the proposed PEBDSAM}
    \label{fig:PEBDSAM}
\end{figure}
Figure~\ref{fig:PEBDSAM} illustrates the core module of the proposed PEBDSAM. The input $\mathbf{x}$ ($\mathbf{x}\in \mathbb{R}^{N \times d}$, where $N$ is the number of nodes and $d$ is the original dimension.) first yields $\text{offset\_num}$ ($M$) offsets via Eq.~\eqref{eq:generate_offset}.
    \begin{align}
    \mathbf{h}_i &= \text{MLP}_{\text{embed}}(\mathbf{x}_i) 
                   = \mathbf{W}_{\text{embed}} \mathbf{x}_i + \mathbf{b}_{\text{embed}} 
                   \in \mathbb{R}^{d_h} \nonumber \\
    \Delta_{i,j} &= \sigma\bigl( \text{MLP}( \mathbf{h}_i \parallel \mathbf{PE}_i ) \bigr), \quad j=1,\dots,M
    \label{eq:generate_offset}
    \end{align}
where $M$ denotes the number of offsets, $\|$ is vector concatenation, and $\text{MLP}$ is implemented as a two-layer fully-connected network with ReLU activation after the first layer, and $\sigma$ denotes the activation function, for which we adopt Tanh in this work.

Then,Eq.~\eqref{eq:obtain_query} is used to compute the $M$ query points. 
\begin{equation}
    \mathbf{q}_{i,j} = \mathbf{PE}_i + \lambda\Delta_{i,j}, \quad j=1,\dots,M
    \label{eq:obtain_query}
\end{equation}
For each query point $\mathbf{q}_{i,j}$, Eq.~\eqref{eq:search pe} is used to locate $K$ relevant nodes in PE space based on Euclidean distance.
\begin{equation}
    \left\{ \left( \mathbf{PE}_{i,j}^k, \mathbf{fea}_{i,j}^k \right) \right\}_{k=1}^{K}
=
\text{Nearest}_K \left( \mathbf{q}_{i,j} ; \mathcal{H} \right), \quad j=1,\dots,M, \quad k=1,\dots,K
    \label{eq:search pe}
\end{equation}
where \(\mathcal{H}\) denotes the positional encoding (PE) space spanned by all nodes in the graph, \(\mathbf{PE}_{i,j}^k\) and \(\mathbf{fea}_{i,j}^k\) are the positional encoding and feature of the \(k\)-th nearest node to the query point \(\mathbf{q}_{i,j}\) in \(\mathcal{H}\), respectively, and \(K\) is the number of retrieved neighbors.

\begin{figure}[htb]
    \centering
    % 这里 your_image 不要带后缀名（如果有多个后缀，会按 eps/pdf/png 顺序匹配）
    \includegraphics[width=0.8\columnwidth]{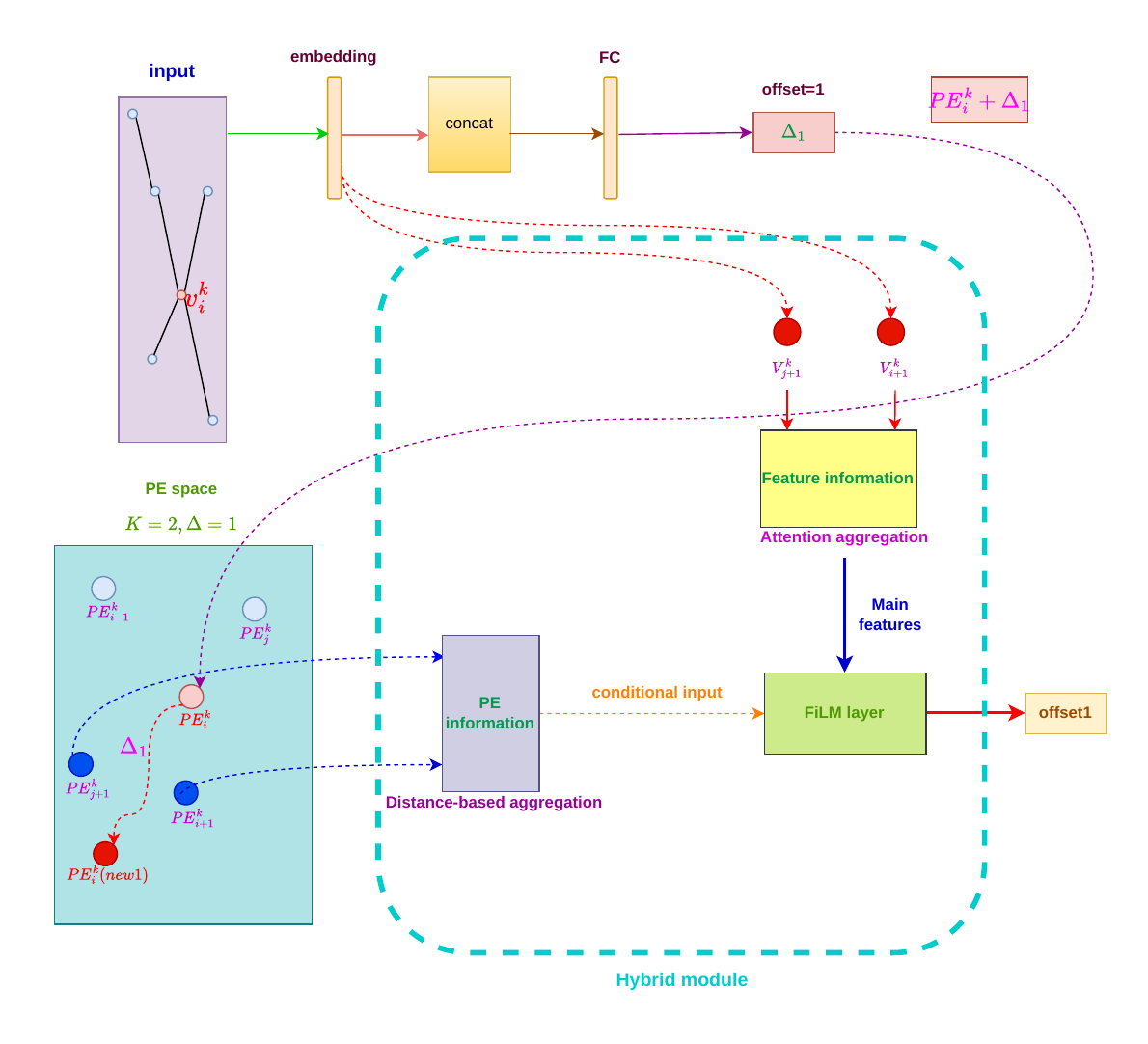}
    \caption{Illustration of the proposed hybrid module, we employ $\text{FiLM}$ layers to achieve this.}
    \label{fig:FILM}
\end{figure}

Now, we obtain the feature information (\(\mathbf{fea}_{i,j}^k\)) and the position information (\(\mathbf{PE}_{i,j}^k\)) of these \(K\) nodes via Eq.~\eqref{eq:search pe}. Next, the information of these \(K\) points is fused using a hybrid module (See Fig.~\ref{fig:FILM} for a detailed illustration.). 

Currently, in the field of GNNs, various methods can be used to fuse the structural and feature information. For instance, \citet{dwivedi2021generalization} proposed a concatenation operation, whereas \citet{dwivedi2021generalization,zhang2021graphbert} et al. introduced element‑wise addition. In addition, \citet{ying2021neurips-transformers} proposed the use of positional information as an attention bias. \citet{2021arXiv210605667M} formulates graph structural information as a multiplicative modulation on attention logits, in stark contrast to the additive bias or input feature concatenation employed by prior mainstream architectures. \citet{perez2018film} introduced the FiLM module. For simplicity in this model's design, we utilize the FiLM module as the fusion module, with its detailed structure shown in Figure~\ref{fig:FILM}.
\begin{align}
    \mathbf{\text{PE-Agg}}_{i,j} &= \sum_{k=1}^{K} w_{i,j}^k \, \mathbf{PE}_{i,j}^{k}, \quad j=1,\dots,M. \nonumber \\ 
    \alpha_{i,j}^k &=
\frac{
\exp\left( \text{LeakyReLU}\left( \mathbf{a}^{\top} \left[ \mathbf{W}_C\,\mathbf{h}_i \;\Vert\; \mathbf{W}_N\,\mathbf{\text{fea}}_{i,j}^{k} \right] \right) \right)
}{
\sum_{k=1}^{K} \exp\left( \text{LeakyReLU}\left( \mathbf{a}^{\top} \left[ \mathbf{W}_C\,\mathbf{h}_i \;\Vert\; \mathbf{W}_N\,\mathbf{\text{fea}}_{i,j}^{k} \right] \right) \right)
}. \nonumber \\
    \mathbf{\text{F-Agg}}_{i,j} &= \sum_{k=1}^{K} \alpha_{i,j}^k \, \mathbf{W}_N \, \mathbf{\text{fea}}_{i,j}^{k}, \quad j=1,\dots,M.
    \label{eq:agg_fea}
\end{align}
where $\mathbf{PE}_{i,j}^{k}$ denotes the Laplacian positional encoding of the $k$-th selected neighbor, and $w_{i,j}^k$ is the weight obtained by applying a $\text{Softmax}$ function to the similarity between the query position and the positions of the neighboring nodes;$\mathbf{W}_C,\mathbf{W}_N$ stand for learnable parameter matrices; $\mathbf{a}$ is the attention‑parameter vector; $\mathbf{\text{fea}}_{i,j}^{k} \,and\, \mathbf{PE}_{i,j}^{k}$ refer to the feature and positional encoding of the $k$-th neighbor, respectively.

We first obtain the aggregated position information $(\mathbf{\text{PE-Agg}}_{i,j})$ and corresponding features $(\mathbf{\text{F-Agg}}_{i,j})$ of k spatial nodes via Eq.~\eqref{eq:agg_fea}. The FILM layer is then utilized to modulate structural and spatial information (See Eq.~\ref{eq:film} for detailed formulations.).
\begin{align}
\label{eq:film}
\bm{\gamma}_{i,j} &= \mathbf{W}_\gamma \, \mathbf{\text{PE-Agg}}_{i,j} + \mathbf{b}_\gamma, \nonumber \\
\bm{\beta}_{i,j}  &= \mathbf{W}_\beta \, \mathbf{\text{PE-Agg}}_{i,j} + \mathbf{b}_\beta, \nonumber \\
\mathbf{\text{offset}_{i,j}} &= \bm{\gamma}_{i,j} \odot \mathbf{\text{F-Agg}}_{i,j} + \bm{\beta}_{i,j}, \quad j=1,\dots,M.
\end{align}
where $\mathbf{W}_\gamma, \mathbf{W}_\beta$ and $\mathbf{b}_\gamma, \mathbf{b}_\beta$ are learnable parameters, and $\odot$ denotes the element‑wise product (i.e., the Hadamard product). We initialize $\mathbf{b}_\gamma$ as all‑ones and $\mathbf{b}_\beta$ as all‑zeros at initialization. This setting enables FiLM to behave as an identity mapping initially and facilitates stable training.

At this point, we have $M$ offset features ($\mathbf{\text{offset}_{i,j}}$), which were aggregated using the attention mechanism in Eq.~\eqref{eq:cross_attention} to combine these $M$ offset features.
\begin{align}
\alpha_{i,j} &= \text{Softmax}\frac{ (\mathbf{h}_i \mathbf{W}_Q) \; \cdot \;  (\mathbf{\text{offset}_{i,j}} \mathbf{W}_K) }{ \sqrt{d} } 
 \nonumber \\
\mathbf{V}_{i}^{\text{spa}} &= \sum_{j=1}^{M} \alpha_{i,j} \; \mathbf{\text{offset}_{i,j}}
\label{eq:cross_attention}
\end{align}
where $\mathbf{W}_Q$ and $\mathbf{W}_K$ are learnable parameters, and $d$ denote the dimension of the query and key vectors.

Finally, the spatially‑aware output $\mathbf{V}_{i}^{\text{spa}}$ of PEBDSAM and the structure‑aware output $\mathbf{V}_{i}^{\text{str}}$ yielded by conventional GNNs (See Eq.~\ref{eq:GNN} for detailed formulations)  are fused through the adaptive gating mechanism given in Eq.~\eqref{eq:gate}
\begin{equation}
    \mathbf{V}_{i}^{\text{str}} = \mathbf{\text{GNN}}(\mathbf{h}_i)
    \label{eq:GNN}
\end{equation}
where various graph neural network models can be adopted for the GNN module.

\begin{align}
    \mathbf{g}_i &= \sigma\Big( \mathbf{W}_g \big[ \mathbf{V}_{i}^{\text{spa}} \; \| \; \mathbf{V}_{i}^{\text{str}} \big] + \mathbf{b}_g \Big), \nonumber \\
    \mathbf{h}_i^{\text{fused}} &= \mathbf{g}_i \odot \mathbf{V}_{i}^{\text{str}} + (\mathbf{1} - \mathbf{g}_i) \odot \mathbf{V}_{i}^{\text{spa}}
    \label{eq:gate}
\end{align}
where $\mathbf{W}_g$ and $\mathbf{b}_g$ are learnable parameters, and $\sigma(\cdot)$ denotes the sigmoid activation function; $\odot$ denotes the Hadamard (element-wise) product and $\mathbf{1}$ is an all-ones vector. 
\begin{equation}
\mathbf{h}_i^{\text{l+1}} = \text{ELU}\big( \mathbf{h}_i^{\text{fused(l)}} + \mathbf{h}_i^{l} \big),
    \label{eq:layer_output}    
\end{equation}
We also include residual connections and activation (see Eq~\eqref{eq:layer_output}), where $l$ denotes the $l$-th layer and $l+1$ denotes the $(l+1)$-th layer.

\section{Diagnostic Experiments And Analysis}
\label{sec:dia}     % 给实验打标签

\subsection{Metric Definitions}
\label{sec:dia_defin}
To facilitate analysis and deepen our understanding of the model, we first define several metrics, including their formulas and meanings.

As defined in Eq.~\eqref{eq:offset_mean}, metric $\text{OffsetAbsMean}$ quantifies the absolute magnitude of coordinate offsets learned by the model. If $\text{OffsetAbsMean}$ persistently approaches 1.0 (saturation), the model is driven to exploit the maximum offset, implying that the current $\lambda$ may be too small and constrains its capacity for large‑range displacement. By contrast, if $OffsetAbsMean$ remains low (e.g., \(<0.2\)) while competitive performance is attained, the model only requires fine‑grained local adjustments. In this case, the existing $\lambda$ may already be sufficiently large and could even introduce redundancy.
\begin{equation}
\text{OffsetAbsMean} = \frac{1}{|\Delta_{\mathrm{i,j}}|} \sum_{p} \left| \Delta_{\mathrm{i,j},\,p} \right|
\label{eq:offset_mean}
\end{equation}
where $p$ ranges over all elements of $\Delta_{\mathrm{i,j}}$.

The second metric is defined as $\text{K‑NN Label Homophily}$ (see Eq.~\eqref{eq:k_nn homophily} for details). It quantifies the degree to which the neighbors selected for each node within the positional‑encoding space share the same label as the node itself. This metric facilitates analysis of neighbor‑selection quality within the spatial aggregation branch.
\begin{align}
    D_{ij} &= \lVert \mathbf{PE}_i - \mathbf{PE}_j \rVert_2^2 \nonumber \\
    \mathcal{N}_K(i) &= \arg\min_{j \in \mathcal{V}\setminus\{i\}}^{(K)} D_{ij} \nonumber \\
    w_{ij} &= \frac{\exp(-D_{ij}/T)}{\sum_{l\in\mathcal{N}_K(i)} \exp(-D_{il}/T)} \nonumber \\
    h_i^{\text{w}} &= \sum_{j \in \mathcal{N}_K(i)} w_{ij} \, \mathbbm{1}[y_j = y_i]  \nonumber \\
    \text{K‑NN Label Homophily} &= \frac{1}{N} \sum_{i\in\mathcal{V}} h_i^{\text{w}} 
\label{eq:k_nn homophily}    
\end{align}
\(\text{K‑NN Label Homophily}\in[0,1]\), where a higher value suggests that spatially proximate nodes within the positional‑encoding space tend to share identical labels.

The third metric is $\text{Node Homophily}$(see Eq.~\eqref{eq:node_homophily} for details)\citep{pei2020iclr-geomgcn}, which measures the fraction of $1$‑hop neighbors of a given node that share the same ground‑truth label as the node itself.
\begin{equation}
\text{Node Homophily} = \frac{1}{N}\sum_{i=1}^{N}
\frac{\big|\{j\in \mathcal{N}(i)\,\big|\, y_i = y_j\}\big|}{\big|\mathcal{N}(i)\big|}
\label{eq:node_homophily}
\end{equation}

The fourth metric is the $\text{Overlap Ratio}$, whose definition is provided in Eq.~\eqref{eq:overlap}.
\begin{equation}
\text{Overlap Ratio}(v) = \frac{|\mathcal{N}_{\mathrm{Spa}}(v) \cap \mathcal{N}_{\mathrm{Str}}(v)|}{k}
\label{eq:overlap}
\end{equation}
where $\mathcal{N}_{\mathrm{Str}}$ denotes the set of $1$‑hop neighbors, and $\mathcal{N}_{\mathrm{Spa}}(v)$ represents the set of k nearest neighbors derived from spatial information.

\subsection{Experimental Findings}
When we conducted diagnostic analysis on our model, we observed several phenomena.

\begin{figure}[htb]
    \centering
    % 这里 your_image 不要带后缀名（如果有多个后缀，会按 eps/pdf/png 顺序匹配）
    \includegraphics[width=\columnwidth]{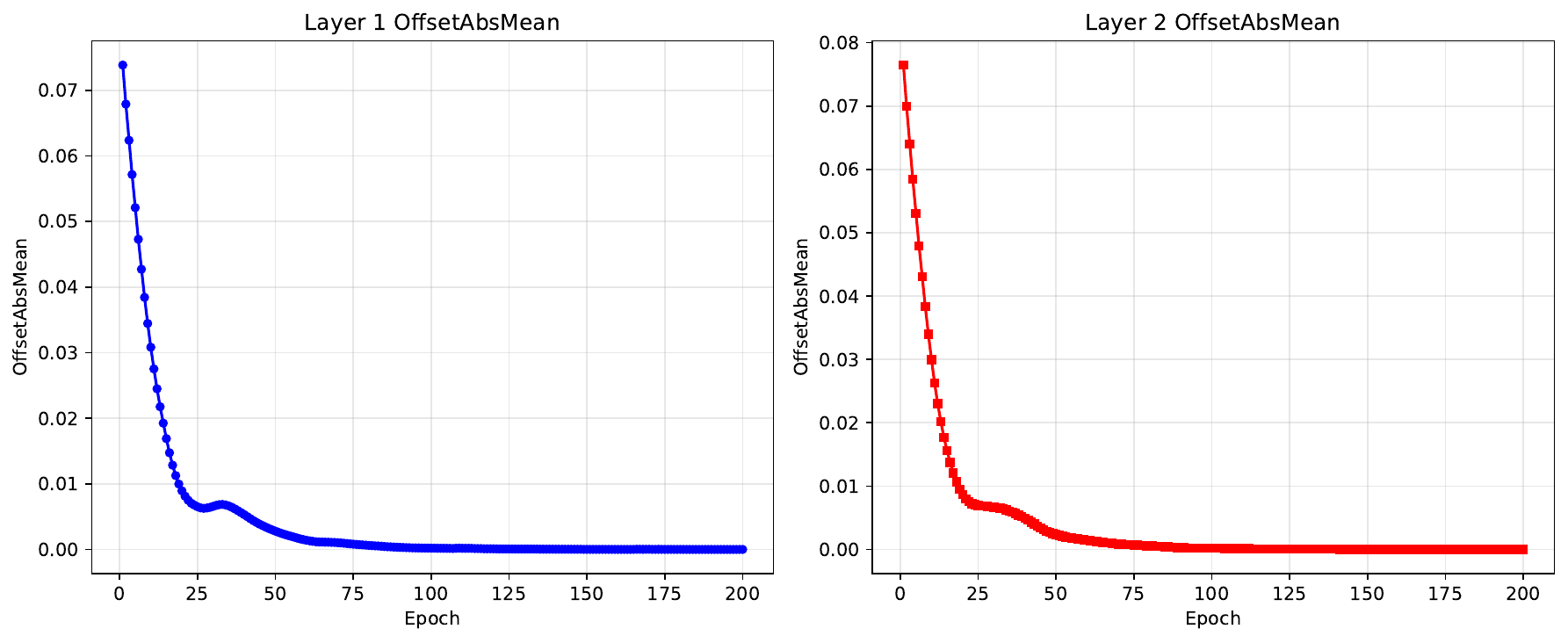}
    \caption{Trends of $\text{OffsetAbsMean}$ across two layers of the model with varying training epochs}
    \label{fig:offset_mean}
\end{figure}
Finding 1: When we trained the model and printed $\text{OffsetAbsMean}$ (Eq.~\eqref{eq:offset_mean}), we found that its value approached 0 as the model training epochs increased (see Figure~\ref{fig:offset_mean}).This indicates that the offset did not play a real role in this experiment.

Finding 2: Since the offset itself played no role, we set $\lambda$ = 0. When we changed $\text{offset\_num}$ ($M$) (e.g., 8→4→1), we observed a decrease in performance (see Table~\ref{tab:M}). 
\begin{table}[ht]
\centering
\caption{Effects of different values of $M$ on node‑classification accuracy over the Chameleon dataset and use Laplacian positional encoding.}
\label{tab:M}
\begin{tabular}{l c c c}
\toprule[1.5pt]   % 顶线更粗（需要加载 booktabs 宏包）
\multicolumn{1}{c}{\textbf{$M$}} & \textbf{1} & \textbf{4} & \textbf{8}  \\
\midrule           % 中线
Accuracy (\%) & $70.60 \pm 0.24$ & $71.93 \pm 0.34$ & $72.41 \pm 0.39$ \\
\bottomrule[1.5pt] % 底线更粗
\end{tabular}
\end{table}
However, in theory, when $\lambda$ is set to 0, the final results should be independent of the quantity of $M$ ($M > 1$). This is because when the $\lambda = 0$, each query  is identical (see Eq.~\eqref{eq:obtain_query}), so every $\mathbf{q}_{i,j}$ would select exactly the same k nearest neighbors (see Eq.~\eqref{eq:search pe}). After $\mathbf{\text{PE-Agg}}_{i,j}$ is aggregated through Eq.~\eqref{eq:agg_fea}, in theory, each $\mathbf{\text{PE-Agg}}_{i,j}$($j=1,\dots,M$) remains exactly the same. After modulation by $FiLM$, $\mathbf{\text{offset}_{i,j}}$ is also identical ($j=1,\dots,M$) (see Eq.~\eqref{eq:film}). The subsequent $att_weights$ then degenerate to a uniform distribution (weight for each offset = 1/$offset_num$) (see Eq.~\eqref{eq:cross_attention}). In the end, $\mathbf{V}_{i}^{\text{spa}}$ is simply the average of $\mathbf{\text{offset}_{i,j}}$, which equals the result for a single offset (i.e.,$M = 1$), i.e., $\mathbf{V}_{i}^{\text{spa}}$ is independent of $offset_num$ (M). So the final output out should also be unrelated to $offset_num$.
\subsection{Analysis}
\begin{figure}[htbp]
    \centering
    % 第一行
    \begin{subfigure}[b]{0.45\textwidth}
        \centering
        \includegraphics[width=\linewidth]{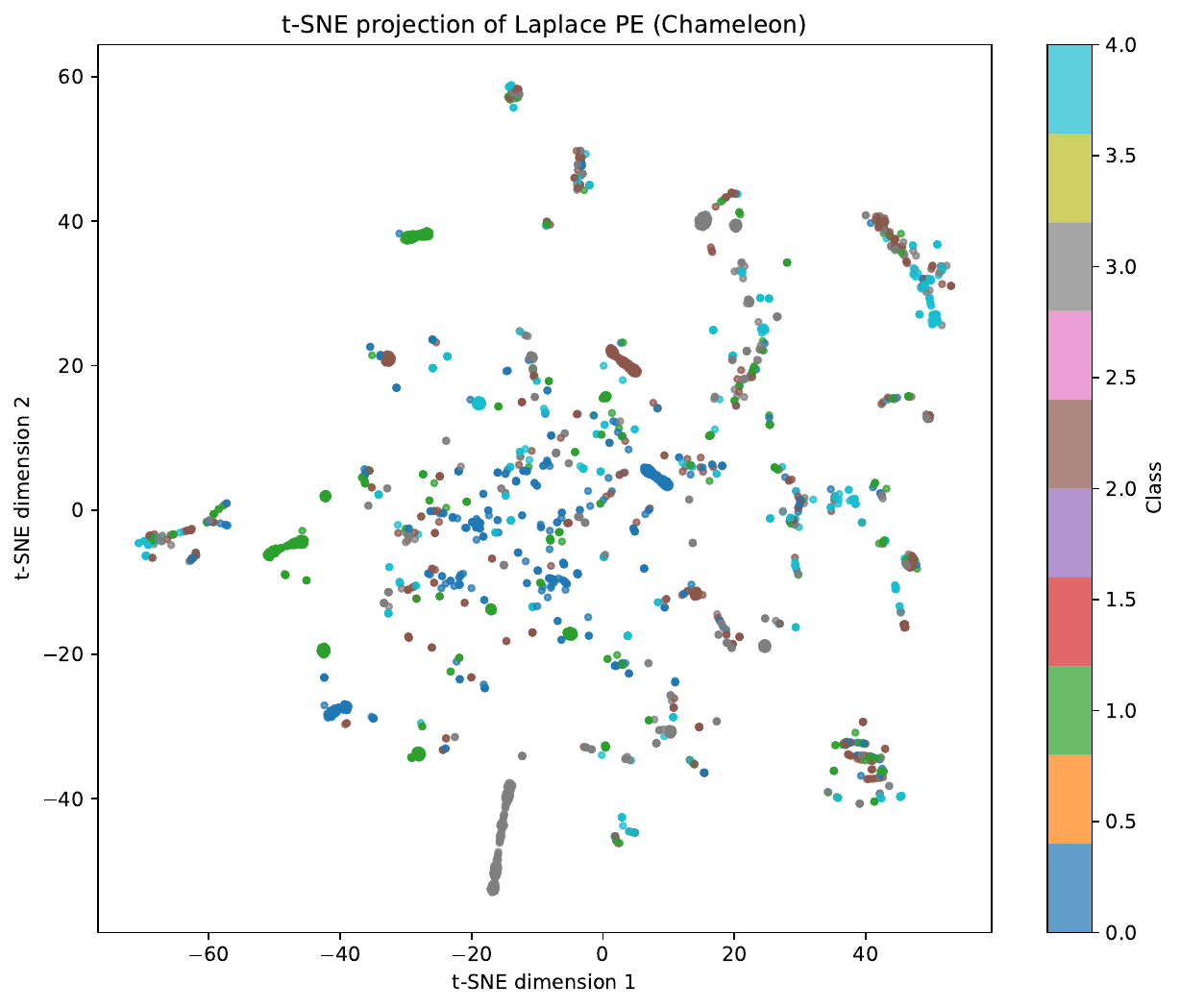}
        \caption{} % 留空则只显示 (a)
        \label{fig:sub1}
    \end{subfigure}
    \hfill % 水平填充间距
    \begin{subfigure}[b]{0.45\textwidth}
        \centering
        \includegraphics[width=\linewidth]{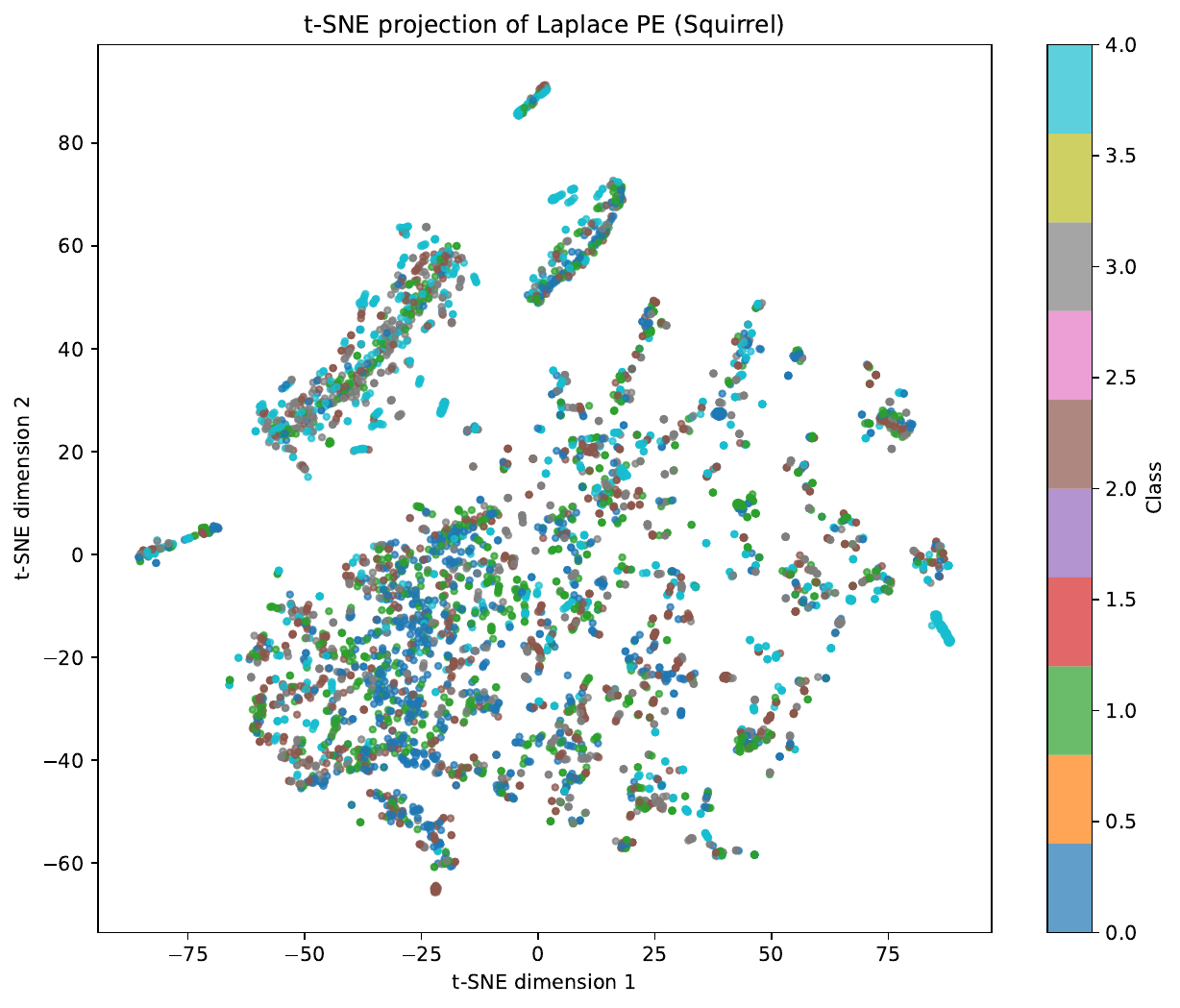}
        \caption{}
        \label{fig:sub2}
    \end{subfigure}
    \\ % 换行
    \bigskip % 垂直间距
    % 第二行
    \begin{subfigure}[b]{0.45\textwidth}
        \centering
        \includegraphics[width=\linewidth]{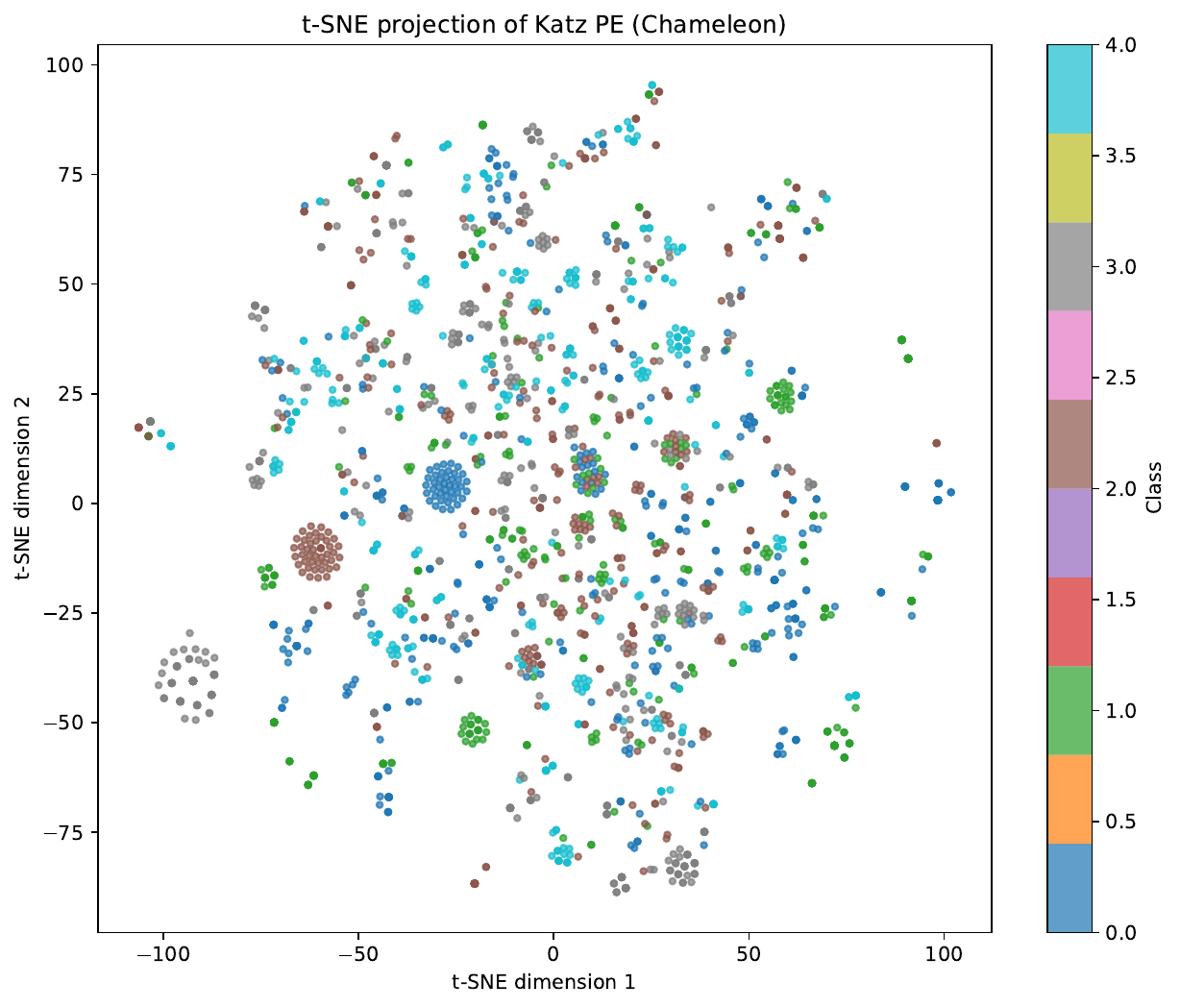}
        \caption{}
        \label{fig:sub3}
    \end{subfigure}
    \hfill
    \begin{subfigure}[b]{0.45\textwidth}
        \centering
        \includegraphics[width=\linewidth]{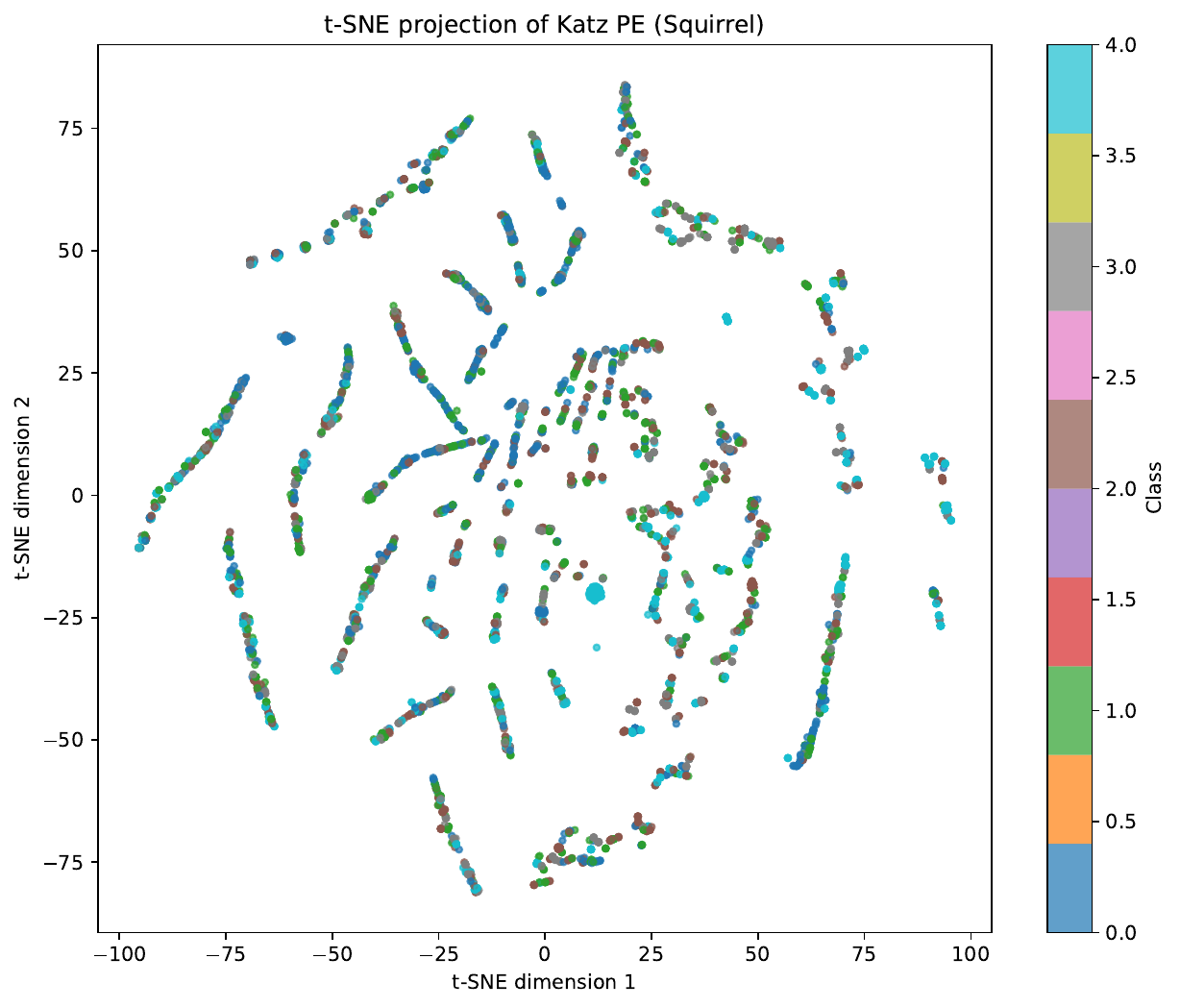}
        \caption{}
        \label{fig:sub4}
    \end{subfigure}
    % 全局标题（4张图共用一个）
    \caption{$2D$ t‑SNE visualization of Laplacian positional encoding and $Katz$ positional encoding on the Chameleon and Squirrel datasets,where Colors indicate ground‑truth node labels.}

    \label{fig:t_sne}
\end{figure}

We conducted a detailed analysis of the first phenomenon. Figure~\ref{fig:t_sne} shows the $2D$ visualization of Laplacian and Katz positional encodings on the Chameleon and Squirrel datasets. As shown in Figure~\ref{fig:t_sne}, points with the same color represent the same labels, and we can see that points with the same label do not cluster together but are rather scattered. This implies that for the query points with the added offset calculated by Formula~\eqref{eq:obtain_query}, most of the surrounding points are noise. Since the distribution of these points is nearly random, the deformable mechanism cannot learn meaningful offsets. This causes the offset ($\Delta_{i,j}$) to trend toward zero and stops further exploration outward. This indicates that the deformable offset is completely degenerated, and the model degenerates into fixed PE nearest neighbor aggregation. In the future, it will be necessary to design more advanced and suitable positional encodings to resolve this issue.

For the second phenomenon, we carefully reviewed the logic in the code. We observe that Dropout is applied after computing the attention coefficient $\alpha_{i,j}^k$ in Eq.~\eqref{eq:agg_fea}. We suspected that Dropout might have affected the final experimental results. Accordingly, we eliminate dropout following the computation of attention coefficient $\alpha_{i,j}^k$, and found that the final results were almost unchanged (see Table~\ref{tab:M-remove-dropout}). This indicates that although the multi-offset design does not produce meaningful dynamic jumps, it forms a cheap yet effective random ensemble through dropout, improving generalization.
\begin{table}[ht]
\centering
\caption{Effects of different values of $M$ on node‑classification accuracy over the Chameleon dataset, we omit dropout after $\alpha_{i,j}^k$ and use Laplacian positional encoding.}
\label{tab:M-remove-dropout}
\begin{tabular}{l c c c}
\toprule[1.5pt]   % 顶线更粗（需要加载 booktabs 宏包）
\multicolumn{1}{c}{\textbf{$M$}} & \textbf{1} & \textbf{4} & \textbf{8}  \\
\midrule           % 中线
Accuracy (\%) & $71.76 \pm 0.13$ & $72.44 \pm 0.33$ & $71.42 \pm 0.35$ \\
\bottomrule[1.5pt] % 底线更粗
\end{tabular}
\end{table}
\subsection{In-depth Analysis}

Having analyzed up to this point, a natural question arises: since the deformable offset mechanism has become ineffective, why is the model still able to achieve good performance?

\begin{table}[ht]
\centering
\small
\setlength{\tabcolsep}{4pt} % 缩小列之间空隙，默认6pt
\caption{We investigate the effects of different position-encoding types, position-encoding dimensionality, and the number of neighbors k on the latent representation \(\text{LH}\) over the Chameleon dataset. Here, $k$ denotes the number of neighbors, $pe$ denotes the dimensionality of position encoding, and \(\text{LH}\) is the $\text{K-NN Label Homophily}$ defined by Equation~\eqref{eq:k_nn homophily}.}
\label{tab:knn_homo}
\resizebox{\textwidth}{!}{%
\begin{tabular}{l c c c c c c c c c}
\toprule[1.5pt]   % 顶线更粗（需要加载 booktabs 宏包）
\multicolumn{1}{c}{\textbf{Laplace PE}} & \textbf{Katz PE} & \textbf{$pe=2$} & \textbf{$pe=4$} & \textbf{$pe=6$} & \textbf{$pe=8$} & \textbf{$K=2$} & \textbf{$K=4$} & \textbf{$K=6$} & \textbf{LH(\%)} \\
\midrule           % 中线
\multicolumn{1}{c}{$\checkmark$} & $\times$ & $\checkmark$ & $\times$ & $\times$ & $\times$ & $\checkmark$ & $\times$ & $\times$ & 77.10 \\
\midrule           % 中线
\multicolumn{1}{c}{$\checkmark$} & $\times$ & $\times$ & $\checkmark$ & $\times$ & $\times$ & $\checkmark$ & $\times$ & $\times$ & 79.27 \\
\midrule           % 中线
\multicolumn{1}{c}{$\checkmark$} & $\times$ & $\times$ & $\times$ & $\checkmark$ & $\times$ & $\checkmark$ & $\times$ & $\times$ & 79.69 \\
\midrule           % 中线
\multicolumn{1}{c}{$\checkmark$} & $\times$ & $\times$ & $\times$ & $\times$ & $\checkmark$ & $\checkmark$ & $\times$ & $\times$ & 79.38 \\
\midrule           % 中线
\multicolumn{1}{c}{$\checkmark$} & $\times$ & $\times$ & $\times$ & $\times$ & $\checkmark$ & $\times$ & $\checkmark$ & $\times$ & 72.61 \\
\midrule           % 中线
\multicolumn{1}{c}{$\checkmark$} & $\times$ & $\times$ & $\times$ & $\times$ & $\checkmark$ & $\times$ & $\times$ & $\checkmark$ & 68.38 \\
\midrule           % 中线
\multicolumn{1}{c}{$\times$} & $\checkmark$ & $\times$ & $\times$ & $\times$ & $\checkmark$ & $\checkmark$ & $\times$ & $\times$ & 73.87 \\
\bottomrule[1.5pt] % 底线更粗
\end{tabular}
}
\end{table}
We conducted a quantitative analysis using the $\text{K‑NN Label Homophily}$ metric (Eq.~\eqref{eq:k_nn homophily}) defined in Subsection~\ref{sec:dia_defin}, which measures the extent to which the neighbors selected in the position encoding space share the same label as the node. We computed this metric for Laplace position encoding and $Katz$ position encoding on the Chameleon and Actor datasets, with results shown in Table~\ref{tab:knn_homo} and Table~\ref{tab:knn_homo-actor}. As seen from Table~\ref{tab:knn_homo}, although Laplace position encoding and $Katz$ position encoding do not cluster same-class nodes together, there are still same-class nodes among the K nearest points around the query node, providing useful information. After filtering through the internal attention mechanism of PEBSAM, $V_{spa}$ ultimately contains meaningful information from these same-class nodes. This explains why the model can still see improved performance even after the deformable mechanism fails.

\begin{table}[ht]
\centering
\caption{Correlation between $\text{K‑NN Label Homophily}$ and model performance}
\label{tab:k-nn and model perfection}
\begin{tabular}{l c c c c}
\toprule[1.5pt]   % 顶线更粗（需要加载 booktabs 宏包）
\multicolumn{1}{c}{\textbf{\text{PE-type}}} & \textbf{\text{PE-dim}} & \textbf{K} & \textbf{LH(\%)} & \textbf{Accuracy (\%)}   \\
\midrule           % 中线
\multicolumn{1}{c}{{\text{Katz}}} & 8 & 2 & 73.47 & $71.81 \pm 0.57$  \\
\midrule           % 中线
\multicolumn{1}{c}{{\text{Laplace}}} & 8 & 2 & 79.38 & $71.93 \pm 0.34$  \\
\midrule           % 中线
\multicolumn{1}{c}{{\text{Laplace}}} & 8 & 4 & 72.61 & $71.69 \pm 0.31$  \\
\midrule           % 中线
\multicolumn{1}{c}{{\text{Laplace}}} & 8 & 6 & 68.38 & $68.10 \pm 0.19$  \\
\midrule           % 中线
\multicolumn{1}{c}{{\text{Laplace}}} & 2 & 2 & 77.10 & $70.56 \pm 0.35$  \\
\bottomrule[1.5pt] % 底线更粗
\end{tabular}
\end{table}
To further demonstrate the correlation between the $\text{K‑NN Label Homophily}$ metric and model performance, we conducted additional experiments. On the same dataset, the $\text{K‑NN Label Homophily}$ metric is influenced by different position encodings, the dimension of the position encoding, and the number K of selected neighbors (see Table~\ref{tab:knn_homo}. We carried out experiments on the Chameleon  dataset (see Table~\ref{tab:k-nn and model perfection}). From Table~\ref{tab:k-nn and model perfection}, we observe that different $\text{K‑NN Label Homophily}$ metrics yield different model accuracies; a higher $\text{K‑NN Label Homophily}$ metric provides more meaningful information, and correspondingly, the model performs better.

\begin{table}[ht]
\centering
\small
\setlength{\tabcolsep}{4pt} % 缩小列之间空隙，默认6pt
\caption{We investigate the effects of position‑encoding dimensionality, and the number of neighbors k on the latent representation \(\text{LH}\) over the Actor dataset. Here, $k$ denotes the number of neighbors, $pe$ denotes the dimensionality of position encoding, and \(\text{LH}\) is the $\text{K‑NN Label Homophily}$ defined by Equation~\eqref{eq:k_nn homophily}.}
\label{tab:knn_homo-actor}
\begin{tabular}{l c c c c c c c c}
\toprule[1.5pt]   % 顶线更粗（需要加载 booktabs 宏包）
\multicolumn{1}{c}{\textbf{Laplace PE}} & \textbf{$pe=2$} & \textbf{$pe=4$} & \textbf{$pe=6$} & \textbf{$pe=8$} & \textbf{$K=2$} & \textbf{$K=4$} & \textbf{$K=6$} & \textbf{LH(\%)} \\
\midrule           % 中线
\multicolumn{1}{c}{\checkmark} & $\checkmark$ & $\times$ & $\times$ & $\times$ & $\checkmark$ & $\times$ & $\times$ & 22.10 \\
\midrule           % 中线
\multicolumn{1}{c}{\checkmark} & $\times$ & $\checkmark$ & $\times$ & $\times$ & $\checkmark$ & $\times$ & $\times$ & 20.72 \\
\midrule           % 中线
\multicolumn{1}{c}{\checkmark} & $\times$ & $\times$ & $\checkmark$ & $\times$ & $\checkmark$ & $\times$ & $\times$ & 21.39 \\
\midrule           % 中线
\multicolumn{1}{c}{\checkmark} & $\times$ & $\times$ & $\times$ & $\checkmark$ & $\checkmark$ & $\times$ & $\times$ & 21.28 \\
\midrule           % 中线
\multicolumn{1}{c}{\checkmark} & $\times$ & $\times$ & $\times$ & $\checkmark$ & $\times$ & $\checkmark$ & $\times$ & 20.87 \\
\midrule           % 中线
\multicolumn{1}{c}{\checkmark} & $\times$ & $\times$ & $\times$ & $\checkmark$ & $\times$ & $\times$ & $\checkmark$ & 21.00 \\
\bottomrule[1.5pt] % 底线更粗
\end{tabular}
\end{table}
However, a closer examination of the data in Table~\ref{tab:knn_homo-actor} raises a new problem. On the Actor dataset, the $\text{K‑NN Label Homophily}$ metric is very low, only around 20\% (For the Actor dataset, there are only 5 categories in total, so if neighbors are chosen completely at random, the expected homophily is about 1/5 = 0.2. The results above are barely different from this random baseline. This is likely related to the current position encoding, and learning-based or better-performing position encodings are needed to address this issue.). In terms of structural information $V_{Str}$, according to \citet{pei2020iclr-geomgcn}, the $\text{Node Homophily}$ is $0.20 \to 0.22$. This gives rise to a new question: In $V_{Str}$, structural information contains only 20\% valid information, and spatial information likewise accounts for merely 20\% valid information. If gated fusion (e.g., simple averaging) is adopted for aggregation, does the resultant representation retain merely 20\% valid information in the end? Does this imply that structural information alone is sufficient and that adding complex spatial information does not improve model performance? Is the spatial module redundant? To answer this, we need the $\text{Overlap Ratio}(v)$ metric proposed in Subsection~\ref{sec:dia_defin}. In fact, the key question is whether structural information and spatial information are redundant. 
\begin{table}[ht]
\centering
\caption{$\text{Overlap Ratio}(v)$ under $pe_{dim} = 2$ on the Actor Dataset, where $\mathcal{N}$ denotes the set of $first-order$ neighbors in the original graph, and \textbf{$\mathcal{PE}$} denotes the set of K‑nearest neighbors in the PE space.}
\label{tab:overlap}
\begin{tabular}{l c c c}
\toprule[1.5pt]   % 顶线更粗（需要加载 booktabs 宏包）
\multicolumn{1}{c}{\textbf{$Node$}} & \textbf{$\mathcal{N}$} & \textbf{$\mathcal{PE}$} & \textbf{Overlap Ratio(\%)}  \\
\midrule           % 中线
Node:0 & [812, 2051, 6341] & [5930, 3912] & 0.0  \\
\midrule           % 中线
Node:1 & [3809] & [3196, 6128] & 0.0  \\
\midrule           % 中线
Node:2 &  [93, 258, 361 ... 7571] & [1437, 1714] & 0.0  \\
\midrule           % 中线
All Node &  $\backslash$ & $\backslash$ & 0.86  \\
\bottomrule[1.5pt] % 底线更粗
\end{tabular}
\end{table}
From Table~\ref{tab:overlap}, for the Actor dataset, taking nodes 0, 1, and 2 as columns, the effective information in the structural feature $V_{Str}$ and the effective information in the spatial feature $V_{Spa}$ do not overlap. This means that after filtering via attention in structural and spatial branches and selection via the gating mechanism, the final output contains more than 20\% effective information (e.g., possibly 30\%). The ablation study of the gating mechanism in Section~\ref{subsec:ablation} further confirms this point.

\section{Simplified Model}
\label{sec:sim}     % 给实验打标签

\begin{figure}[htb]
    \centering
    % 这里 your_image 不要带后缀名（如果有多个后缀，会按 eps/pdf/png 顺序匹配）Illustration of the proposed PEBDSAM
    \includegraphics[width=\columnwidth]{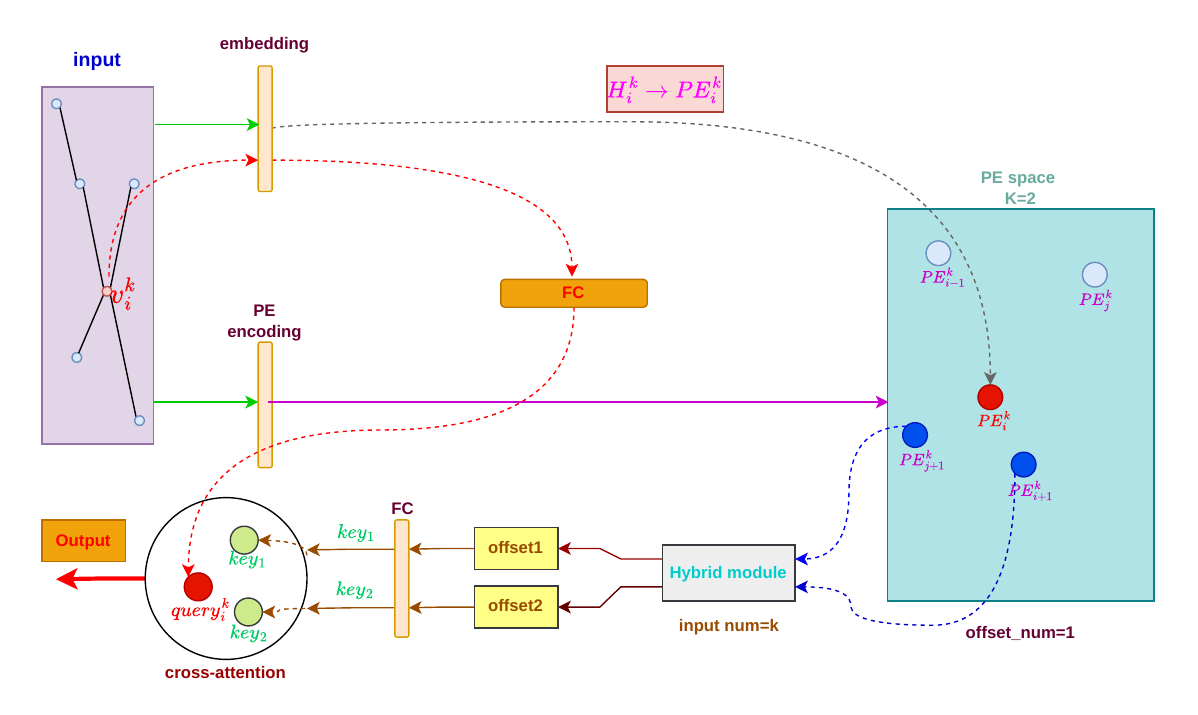}
    \caption{Illustration of the proposed PEBSAM}
    \label{fig:PEBSAM}
\end{figure}
Through the above diagnostic experiments, since the offset approaches zero during training and thus plays virtually no role, we can simply remove the offset branch to reduce the number of parameters, resulting in a streamlined model. The PEBDSAM after removing the offset is referred to as the PEBSAM. The structure of the PEBSAM is shown in Figure~\ref{fig:PEBSAM}. The overall workflow is similar to that in Figure~\ref{fig:PEBDSAM}, except that the offset module has been removed. By removing $offset_{mlp}$ (i.e., eliminating $\lambda\Delta_{i,j}$ in Eq.~\eqref{eq:obtain_query} and $\Delta_{i,j}$ in Eq.~\eqref{eq:generate_offset}), each node directly selects the k-nearest neighbors in the PE space.

\section{Experiments}
\label{sec:exp}     % 给实验打标签

In this section, to verify the generality and effectiveness of our framework, we conducted experiments on three homophilic graphs and six heterophilic graphs; these datasets pertain to node classification tasks.

\subsection{Datasets}

We selected three homophilic graphs and six heterophilic graphs for experimentation. All nine datasets are publicly available and can be directly loaded using the PyTorch Geometric (PyG) library \citep{Fey/Lenssen/2019}. We adopted the splits proposed in the Geom-GCN paper \citep{pei2020iclr-geomgcn}, in which each dataset provides 10 predefined public splits. This facilitates both reproducibility and fairness. Table~\ref{tab:dataset_stats} provides the statistics for these datasets.
\begin{table}[htbp]
    \centering
    \caption{Dataset statistics. \cite{pei2020iclr-geomgcn}}
    \label{tab:dataset_stats}
    \begin{tabular}{l|ccccccccc}
        \hline
        Dataset & Cora & Cite. & Pubm. & Cham. & Squi. & Actor & Corn. & Texa. & Wisc. \\
        \hline
        Nodes & 2708 & 3327 & 19717 & 2277 & 5201 & 7600 & 183 & 183 & 251 \\
        Edges & 5429 & 4732 & 44338 & 36101 & 217073 & 33544 & 295 & 309 & 499 \\
        Features & 1433 & 3703 & 500 & 2325 & 2089 & 931 & 1703 & 1703 & 1703 \\
        Classes & 7 & 6 & 3 & 5 & 5 & 5 & 5 & 5 & 5 \\
        \hline
    \end{tabular}
\end{table}

\subsection{Baselines and Experimental Settings}

Baselines: For the baseline models, we selected traditional GNN models, namely, GCN, GAT, GIN, and GraphSAGE. These models were integrated into the PyG library, enabling convenient loading and use.All baseline models are built with a two‑layer architecture. For GAT, we set the number of attention heads to one and employ ELU \citep{clevert2016fast} as the activation function between layers. For GCN, ReLU \citep{nair2010rectified} is adopted as the layer‑wise activation function. For GIN, we use sum aggregation with a two‑layer MLP as the update function and ReLU activation, and fix $\epsilon=0$. For GraphSAGE, we apply the mean aggregator and use ReLU as the layer‑wise activation function.

Experimental settings: All models used two layers, the embedding layer dimension was set uniformly to 64, the number of k-nearest neighbors was set uniformly to 2, dropout was set uniformly to 0.5, and the learning rate was set uniformly to 0.005. We set the maximum number of epochs to 1000 and the patience of early stopping to 100 for all experiments. The optimizer was uniformly set to Adam \citep{kingma2015adam}, and weight-decay was set uniformly to 5e-4. The loss function is uniformly set to the cross-entropy classification loss \citep{bridle1990probabilistic} and each dataset is independently run for ten repeated trials. For the Chameleon, Squirrel, Texas, Wisconsin and Cornell datasets, $offset-num$ was set to 16; for the Actor, Cora, Pubmed and Citeseer datasets, $offset-num$ was set to 8. For the Pubmed dataset, we set the position encoding dimension to 1 and used the PEBSAM-Speed; for other datasets, we used the PEBSAM and uniformly set the position encoding dimension to 8.

\subsection{Experimental Results}

\begin{table}[htbp]
    \centering
    \small
    \setlength{\tabcolsep}{5pt}
    \caption{Model accuracy for each dataset, with Laplacian positional encoding adopted for position encoding. + denotes embedding our plug-and-play module (PEBSAM or PEBSAM-Speed); for the Pubmed dataset, + refers to the PEBSAM-Speed, while for other datasets, + refers to the PEBSAM. The final results are reported as the mean ± standard deviation across 10 independent experiments.}
    \label{tab:node_classification}
    \resizebox{\textwidth}{!}{%
    \begin{tabular}{l|cccccccc}
        \toprule
        Dataset & GCN & GCN+ & GAT & GAT+ & GIN & GIN+ & GraphSAGE & GraphSAGE+  \\
        \midrule
        Cora          & 85.89$\pm$0.23 & 85.50$\pm$0.24 & 86.53$\pm$0.21 & \textbf{86.84$\pm$0.26} & 84.24$\pm$0.35 & 85.56$\pm$0.32 & 84.79$\pm$0.26 & 85.22$\pm$0.34  \\
        Citeseer          & 75.52$\pm$0.26 & 74.24$\pm$0.18 & \textbf{75.57$\pm$0.14} & 73.94$\pm$0.19 & 73.17$\pm$0.34 & 74.57$\pm$0.38 & 74.54$\pm$0.21 & 74.90$\pm$0.24 \\
        Pubmed          & 88.10$\pm$0.05 & 88.50$\pm$0.05 & 86.84$\pm$0.07 & 88.43$\pm$0.04 & 88.35$\pm$0.06 & 88.33$\pm$0.06 & 88.76$\pm$0.05 & \textbf{88.78$\pm$0.05}  \\
        Chameleon     & 38.93$\pm$0.46 & 71.14$\pm$0.27 & 44.39$\pm$0.64 & \textbf{72.38$\pm$0.42} & 39.81$\pm$0.63 & 71.93$\pm$0.26 & 48.79$\pm$0.42 & 70.68$\pm$0.37  \\
        Squirrel        & 27.73$\pm$0.42 & 56.02$\pm$0.33 & 30.09$\pm$0.32 & 51.17$\pm$0.30 & 31.85$\pm$0.39 & \textbf{57.12$\pm$0.64} & 36.48$\pm$0.43 & 53.79$\pm$0.22  \\
        Actor       & 28.51$\pm$0.19 & 33.77$\pm$0.15 & 28.62$\pm$0.21 & 34.54$\pm$0.15 & 28.87$\pm$0.27 & 34.10$\pm$0.27 & 35.64$\pm$0.30 & \textbf{35.94$\pm$0.22}  \\
        Texas     & 59.32$\pm$1.05 & 68.49$\pm$0.74 & 59.46$\pm$0.85 & 71.14$\pm$2.01 & 59.95$\pm$0.95 & 69.16$\pm$1.43 & 78.97$\pm$1.20 & \textbf{80.24$\pm$1.21}  \\
        Wisconsin     & 52.65$\pm$1.16 & 70.76$\pm$0.90 & 52.39$\pm$1.09 & 73.08$\pm$1.30 & 53.67$\pm$1.16 & 73.84$\pm$1.30 & 82.37$\pm$0.98 & \textbf{83.12$\pm$1.14}  \\
        Cornell     & 42.46$\pm$0.99 & 58.84$\pm$1.49 & 43.43$\pm$0.75 & 60.89$\pm$1.80 & 49.49$\pm$0.84 & 62.16$\pm$2.05 & \textbf{73.59$\pm$0.77}  & 70.86$\pm$0.97  \\
        \bottomrule
    \end{tabular}
    }
\end{table}
The experimental results are shown in Table~\ref{tab:node_classification}. As can be seen from Table~\ref{tab:node_classification}, on heterophilic graphs, our PEBSAM consistently achieves stable performance improvements; on homophilic graphs, our PEBSAM also achieves slight improvements, but there is a trend of performance decline in some cases—for example, the GAT+ model on the Citeseer dataset performs 1.63\% worse than the GAT model. The specific reasons why our PEBSAM yields less improvement on homophilic graphs compared to heterophilic graphs, and why it sometimes even leads to a decline, are analyzed in detail in Section~\ref{sec:discussion}.

\subsection{Ablation Experiments}
\label{subsec:ablation}

Table~\ref{tab:ablation-pe} presents the ablation study on position encoding. Position encoding is extremely important in our experiments—a well-designed position encoding enables our original PEBDSAM to function as intended and realize its potential. In PEBSAM, it can provide more effective information, ultimately improving the model's accuracy. In the Table~\ref{tab:ablation-pe}, katz-err refers to the results obtained when we initially computed the katz position encoding, setting lambda-max = 1 due to a limited understanding of the Katz calculation principle. This setting leads to imprecise, even erroneous position encoding results. However, when we evaluated Katz-err's $\text{K‑NN Label Homophily}$, we found its value was much higher than Katz and a substantial improvement in performance is also observed, as shown in Table~\ref{tab:ablation-pe}. In the future, we will attempt learnable position encodings and explore other position encodings that better suited to our model.
\begin{table}[ht]
\centering
\caption{Ablation Study on Positional Encoding with offset-num Fixed at 4 and use GAT+ model}
\label{tab:ablation-pe}
\begin{tabular}{l c c c c}
\toprule[1.5pt]   % 顶线更粗（需要加载 booktabs 宏包）
\multicolumn{1}{c}{\textbf{\text{PE-type}}} & \textbf{\text{PE-dim}} & \textbf{LH(\%)} & \textbf{Dataset} & \textbf{Accuracy (\%)}   \\
\midrule           % 中线
\multicolumn{1}{c}{{\text{Laplace}}} & 8 & 79.38 & Chameleon & $72.38 \pm 0.42$  \\
\midrule           % 中线
\multicolumn{1}{c}{{\text{Katz}}} & 8 & 73.47 & Chameleon & $72.24 \pm 0.22$  \\
\midrule           % 中线
\multicolumn{1}{c}{{\text{Katz-err}}} & 8 & 80.86 & Chameleon & $74.73 \pm 0.39$  \\
\midrule           % 中线
\multicolumn{1}{c}{{\text{Laplace}}} & 8 & 67.22 & Squirrel & $51.17 \pm 0.30$  \\
\midrule           % 中线
\multicolumn{1}{c}{{\text{Katz}}} & 8 & 68.65 & Squirrel & $64.78 \pm 0.20$  \\
\midrule           % 中线
\multicolumn{1}{c}{{\text{Katz-err}}} & 8 & 82.25 & Squirrel & $69.80 \pm 0.18$  \\
\bottomrule[1.5pt] % 底线更粗
\end{tabular}
\end{table}

The FiLM module is used to integrate positional information ($\text{PE-Agg}_{i,j}$) and feature information ($\text{F-Agg}_{i,j}$) (see Eq.~\eqref{eq:agg_fea}). To verify the role of the FiLM module, we conducted ablation experiments for FiLM, as shown in Table~\ref{tab:ablation-film}. "Without FiLM" means that positional information is not integrated, and only feature information is used in subsequent computations.
\begin{table}[ht]
\centering
\caption{An ablation study on the FiLM module is performed using the GAT+ model, with offset-num set to 4.}
\label{tab:ablation-film}
\begin{tabular}{l c c c}
\toprule[1.5pt]   % 顶线更粗（需要加载 booktabs 宏包）
\multicolumn{1}{c}{\text{PE-type}} & \textbf{\text{Without FiLM}} & \textbf{Dataset} & \textbf{Accuracy (\%)}   \\
\midrule           % 中线
\multicolumn{1}{c}{{\text{Laplace}}} & No & Chameleon & $72.56 \pm 0.16$  \\
\midrule           % 中线
\multicolumn{1}{c}{{\text{Laplace}}} & Yes & Chameleon & $70.79 \pm 0.23$  \\
\midrule           % 中线
\multicolumn{1}{c}{{\text{Laplace}}} & No & Squirrel & $59.47 \pm 0.17$  \\
\midrule           % 中线
\multicolumn{1}{c}{{\text{Laplace}}} & Yes & Squirrel & $54.16 \pm 0.22$  \\
\bottomrule[1.5pt] % 底线更粗
\end{tabular}
\end{table}

To validate the effectiveness of the gating mechanism and the two-branch information, we performed ablation experiments on gating mechanism, as shown in Table~\ref{tab:ablation-gate}.Because on the Actor dataset, the $\text{K-NN Label Homophily}$ with Katz positional encoding is 20.41\%, the information provided by $V_{Spa}$ is relatively limited, resulting in only a modest improvement in $V_{Str}$ when spatial information is incorporated. However, on the Texas dataset, the $\text{K-NN Label Homophily}$ with Katz positional encoding is 59.51\%, so the proportion of informative content in $V_{Spa}$ is higher, leading to a substantial improvement in $V_{Str}$ when spatial information is incorporated.
\begin{table}[ht]
\centering
\caption{Gating ablation study of GAT+ with Katz positional encoding on the Actor and Texas datasets, with offset-num set to 4. A single $\checkmark$ indicates the single-branch setting, while two $\checkmark$$\checkmark$ indicate the dual-branch setting with gated fusion.}
\label{tab:ablation-gate}
\begin{tabular}{l c c c}
\toprule[1.5pt]   % 顶线更粗（需要加载 booktabs 宏包）
\multicolumn{1}{c}{\text{$V_{Str}$}} & \textbf{\text{$V_{Spa}$}} & \textbf{Dataset} & \textbf{Accuracy (\%)}   \\
\midrule           % 中线
\multicolumn{1}{c}{$\checkmark$} & $\times$ & Actor & $34.22 \pm 0.24$  \\
\midrule           % 中线
\multicolumn{1}{c}{$\times$} & $\checkmark$ & Actor & $32.32 \pm 0.18$  \\
\midrule           % 中线
\multicolumn{1}{c}{$\checkmark$} & $\checkmark$ & Actor & $35.06 \pm 0.13$  \\
\midrule           % 中线
\multicolumn{1}{c}{$\checkmark$} & $\times$ & Texas & $67.30 \pm 1.08$  \\
\midrule           % 中线
\multicolumn{1}{c}{$\times$} & $\checkmark$ & Texas & $71.35 \pm 0.70$  \\
\midrule           % 中线
\multicolumn{1}{c}{$\checkmark$} & $\checkmark$ & Texas & $72.05 \pm 1.25$  \\
\bottomrule[1.5pt] % 底线更粗
\end{tabular}
\end{table}
\subsection{Supplementary Notes}
We have observed that PyG does not automatically convert original directed graphs to undirected graphs when loading datasets by default. In our preliminary experiments, we overlooked this and mistakenly treated the original directed graph as undirected: when invoking PyG’s built-in Laplacian positional encoding, we set is\_undirected=True. This parameter causes PyG to use the eigsh solver designed for symmetric matrices, which only utilizes the upper (or lower) triangle during computation and assumes matrix symmetry by default, thereby making the resulting topology inconsistent with the original directed graph. Therefore, we further computed Laplacian positional encodings for directed graphs using is\_undirected=False, whereby PyG calls the eigs solver, suitable for asymmetric matrices and preserving complete information from the asymmetric adjacency matrix, which better reflects the original directed structure. The experimental results are shown in Table~\ref{tab:supple}: regardless of whether positional encoding scheme Laplacian PE or Corrected-Laplacian PE is used, the core conclusions of this paper remain unchanged. This is because positional encoding only provides each node with positional information, while our quantitative metric K-NN Label Homophily and Overlap Ratio evaluates the impact of this positional information’s quality on the final experimental results. One more point to note: since ARPACK’s iterative Laplacian solver internally generates random initial vectors for the Lanczos iteration, and by default PyG does not fix the ARPACK initial guess vector v0, the Laplacian positional encodings obtained each time are different, resulting in fluctuation in the K-NN Label Homophily and Overlap Ratio metrics based on Laplacian positional encoding.
\begin{table}[ht]
\centering
\caption{Results of GAT+ equipped with corrected Laplacian positional encoding ($\text{offset-num}=4$) on various datasets.It can be observed that on Squirrel, Laplacian PE and Corrected-Laplacian PE exhibit a marked difference in K-NN Label Homophily, leading to a large gap in the final performance. On the other two datasets, however, the K-NN Label Homophily values of Laplacian PE and Corrected-Laplacian PE are comparable, and the final results remain close.}
\label{tab:supple}
\resizebox{\textwidth}{!}{%
\begin{tabular}{l c c c}
\toprule[1.5pt]   % 顶线更粗（需要加载 booktabs 宏包）
\multicolumn{1}{c}{\textbf{$Dataset$}} & \textbf{PE} & \textbf{\text{K-NN Label Homophily(\%)}} & \textbf{Accuracy(\%)}    \\
\midrule           % 中线
\multicolumn{1}{c}{Chameleon} & Laplacian & 78.02 & $72.80 \pm 0.29$   \\
\midrule           % 中线
\multicolumn{1}{c}{Chameleon} & Corrected-Laplacian & 76.66 & $72.71 \pm 0.21$  \\
\midrule           % 中线
\multicolumn{1}{c}{Squirrel} & Laplacian & 66.11 & $57.52 \pm 0.20$  \\
\midrule           % 中线
\multicolumn{1}{c}{Squirrel} & Corrected-Laplacian & 78.22 & $66.17 \pm 0.33$ \\
\midrule           % 中线
\multicolumn{1}{c}{Actor} & Laplacian & 21.59 & $34.69 \pm 0.26$   \\
\midrule           % 中线
\multicolumn{1}{c}{Actor} & Corrected-Laplacian & 20.82 & $34.84 \pm 0.06$  \\
\bottomrule[1.5pt] % 底线更粗
\end{tabular}
}
\end{table}

\section{Discussion}
\label{sec:discussion}

\begin{table}[ht]
\centering
\caption{Metrics K-NN Label Homophily and Overlap Ratio with Laplacian positional encoding on different datasets}
\label{tab:discuss-idex}
\begin{tabular}{l c c}
\toprule[1.5pt]   % 顶线更粗（需要加载 booktabs 宏包）
\multicolumn{1}{c}{\textbf{$Dataset$}} & \textbf{\text{K-NN Label Homophily(\%)}} & \textbf{Overlap Ratio(\%)}    \\
\midrule           % 中线
\multicolumn{1}{c}{Cora} & 23.59 & 2.51   \\
\midrule           % 中线
\multicolumn{1}{c}{Citeseer} & 53.98 & 14.76  \\
\midrule           % 中线
\multicolumn{1}{c}{Texas} & 50.82 & 10.11  \\
\midrule           % 中线
\multicolumn{1}{c}{Wisconsin} & 45.78 & 14.54   \\
\midrule           % 中线
\multicolumn{1}{c}{Cornell} & 35.95 & 7.38  \\
\bottomrule[1.5pt] % 底线更粗
\end{tabular}
\end{table}
Table~\ref{tab:discuss-idex} presents the $\text{K-NN Label Homophily}$ and $\text{Overlap Ratio}$ metrics on several datasets. As shown in the Table~\ref{tab:discuss-idex}, $\text{K‑NN Label Homophily}$ is generally low, with only 23.59\% on the Cora dataset. However, in the Citeseer dataset, $\text{K‑NN Label Homophily}$ is 53.98\%, which still means that half of the data is noise and that the $\text{Overlap Ratio}$ reaches 14.76\%., indicating severe information redundancy. These two issues are the main reasons why our PEBSAM performs poorly on the Citeseer dataset. For large graphs, such as ogbn-arxiv \citep{hu2020open}, we also experimented with PEBSAM-Speed. Since our PEBSAM-Speed requires pe=1, and on large-scale graphs, the information encoding capacity of pe = 1 is very limited due to the enormous number of nodes, both the $\text{K‑NN Label Homophily}$ and $\text{Overlap Ratio}$ metrics are suboptimal, and the final experimental results are slightly lower than the baseline model. However, computing high-dimensional positional encodings on large graphs incurs a significant cost. This can draw inspiration from ensemble learning \citep{hansen1990neural} in machine learning, such as the concept of random forest \citep{breiman2001random}, where multiple different 1-dimensional positional encodings are used to simplify computation, and mechanisms such as attention \citep{vaswani2017attention} are employed to integrate these results.

In our experiments, we set k to a fixed value of 2; whether this is reasonable remains a question. Increasing k introduces more information, but it also brings in more noise. This issue ultimately requires a more advanced positional encoding to address, in order to improve $\text{K-NN Label Homophily}$. Furthermore, on large graphs, the average node degree is typically high, and setting k=2 provides only limited information. Another question is whether it is appropriate to use the same number of neighbors k for all nodes, since each node's degree is different. In the future, we need to explore adaptive neighbor numbers k that vary for each node.

\section{Conclusion}
\label{sec:concl}   % 给结论打标签
%% Use \subsection commands to start a subsection.
In this study, we propose PEBDSAM. We conducted diagnostic experiments and found that the deformable mechanism did not take effect; therefore, we analyzed the reason and why the model still achieved satisfactory performance after the mechanism failed. Based on this analysis, we present a simplified version (PEBSAM). There are still many aspects of this design that can be improved, and there remains significant room for future advancement. The current design significantly relies on positional encoding. In the future, we will explore positional encoding schemes better suited to our model in order to improve metrics such as K‑NN Label Homophily, enable our deformable mechanism to take effect, provide the $V_{Spa}$ branch with more useful information, and potentially eliminate the $V_{Str}$ branch altogether.

%% The Appendices part is started with the command \appendix;
%% appendix sections are then done as normal sections

%%\appendix
%%\section{Example Appendix Section}
%%\label{app1}

%%Appendix text.

%% For citations use: 
%%       \citet{<label>} ==> Lamport (1994)
%%       \citep{<label>} ==> (Lamport, 1994)
%%
%%Example citation, See \citet{lamport94}.

%% If you have bib database file and want bibtex to generate the
%% bibitems, please use
%%
  \bibliographystyle{elsarticle-harv} 
  \bibliography{cas-refs}

@inproceedings{kipf2017semi,
    title     = {Semi-Supervised Classification with Graph Convolutional Networks},
    author    = {Kipf, Thomas N. and Welling, Max},
    booktitle = {International Conference on Learning Representations (ICLR)},
    year      = {2017}
}

@CONFERENCE{Velickovic2018Graph,
  author    = {Veličković, Petar and Cucurull, Guillem and Casanova, Arantxa and Romero, Adriana and Liò, Pietro and Bengio, Yoshua},
  title     = {Graph Attention Networks},
  booktitle = {International Conference on Learning Representations (ICLR)},
  year      = {2018}
}

@CONFERENCE{Hamilton2017Inductive,
  author    = {Hamilton, William L. and Ying, Rex and Leskovec, Jure},
  title     = {Inductive Representation Learning on Large Graphs},
  booktitle = {Advances in Neural Information Processing Systems (NeurIPS)},
  year      = {2017}
}

@CONFERENCE{Defferrard2016Convolutional,
  author    = {Defferrard, Michaël and Bresson, Xavier and Vandergheynst, Pierre},
  title     = {Convolutional Neural Networks on Graphs with Fast Localized Spectral Filtering},
  booktitle = {Advances in Neural Information Processing Systems (NeurIPS)},
  year      = {2016}
}

@ARTICLE{Bruna2013Spectral,
  author  = {Bruna, Joan and Zaremba, Wojciech and Szlam, Arthur and LeCun, Yann},
  title   = {Spectral Networks and Locally Connected Networks on Graphs},
  journal = {arXiv preprint arXiv:1312.6203},
  year    = {2013}
}

@inProceedings{Dai_2017_ICCV,
    author    = {Dai, Jifeng and Qi, Haozhi and Xiong, Yuwen and Li, Yi and Zhang, Guodong and Hu, Han and Wei, Yichen},
    title     = {Deformable Convolutional Networks},
    booktitle = {Proceedings of the IEEE International Conference on Computer Vision (ICCV)},
    month     = {Oct},
    year      = {2017},
    pages     = {764-773}
}

@inproceedings{zhu2021deformable,
  title     = {Deformable {DETR}: {D}eformable Transformers for End-to-End Object Detection},
  author    = {Zhu, Xizhou and Su, Weijie and Lu, Lewei and Li, Bin and Wang, Xiaogang and Dai, Jifeng},
  booktitle = {International Conference on Learning Representations (ICLR)},
  year      = {2021},
}

@inproceedings{park2022deformable,
  title={Deformable Graph Convolutional Networks},
  author={Park, Jinyoung and Yoo, Sungdong and Park, Jihwan and Kim, Hyunwoo J},
  booktitle={Proceedings of the AAAI Conference on Artificial Intelligence},
  volume={36},
  number={7},
  pages={7949--7956},
  year={2022}
}

@article{park2025deformable,
      title={Deformable Graph Transformer},
      author={Park, Jinyoung and Yun, Seongjun and Park, Hyeonjin and Kang, Jaewoo and Jeong, Jisu and Kim, Kyung Min and Ha, Jung Woo and Kim, Hyunwoo J.},
      journal={IEEE Transactions on Pattern Analysis and Machine Intelligence},
      volume={47},
      number={7},
      pages={5385--5396},
      year={2025},
      doi={10.1109/TPAMI.2025.3550281}
}

@InProceedings{Zhu_2019_CVPR,
    author    = {Zhu, Xizhou and Hu, Han and Lin, Stephen and Dai, Jifeng},
    title     = {Deformable ConvNets V2: More Deformable, Better Results},
    booktitle = {Proceedings of the IEEE/CVF Conference on Computer Vision and Pattern Recognition (CVPR)},
    month     = {June},
    year      = {2019},
    pages     = {9308-9316}
}

@inproceedings{dwivedi2021generalization,
  title     = {A Generalization of Transformer Networks to Graphs},
  author    = {Dwivedi, Vijay Prakash and Bresson, Xavier},
  booktitle = {Proceedings of the AAAI Conference on Artificial Intelligence},
  volume    = {35},
  number    = {8},
  pages     = {7370--7377},
  year      = {2021}
}

@article{zhang2021graphbert,
  author  = {Zhang, Jiawei and Zhang, Haopeng and Xia, Congying and Sun, Li},
  journal = {IEEE Transactions on Neural Networks and Learning Systems},
  title   = {Graph-Bert: Only Attention is Needed for Learning Graph Representations},
  year    = {2021},
  volume  = {32},
  number  = {8},
  pages   = {3556--3568},
  doi     = {10.1109/TNNLS.2020.3048375}
}

@inproceedings{ying2021neurips-transformers,
  title     = {{Do Transformers Really Perform Badly for Graph Representation?}},
  author    = {Ying, Chengxuan and Cai, Tianle and Luo, Shengjie and Zheng, Shuxin and Ke, Guolin and He, Di and Shen, Yanming and Liu, Tie-Yan},
  booktitle = {Neural Information Processing Systems (NeurIPS)},
  year      = {2021},
  url       = {https://mlanthology.org/neurips/2021/ying2021neurips-transformers/}
}

@ARTICLE{2021arXiv210605667M,
  author = {{Mialon}, Gr{\'e}goire and {Chen}, Dexiong and {Selosse}, Margot and {Mairal}, Julien},
  title = "{GraphiT: Encoding Graph Structure in Transformers}",
  journal = {arXiv e-prints},
  year = 2021,
  month = jun,
  eid = {arXiv:2106.05667},
  pages = {arXiv:2106.05667},
  doi = {10.48550/arXiv.2106.05667},
  archivePrefix = {arXiv},
  eprint = {2106.05667},
  primaryClass = {cs.LG},
  adsurl = {https://ui.adsabs.harvard.edu/abs/2021arXiv210605667M}
}

@inproceedings{perez2018film,
  title     = {{FiLM}: Visual Reasoning with a General Conditioning Layer},
  author    = {Perez, Ethan and Strub, Florian and De Vries, Harm and Dumoulin, Vincent and Courville, Aaron},
  booktitle = {Proceedings of the Thirty-Second AAAI Conference on Artificial Intelligence (AAAI)},
  pages     = {3942--3951},
  year      = {2018},
  doi       = {10.1609/aaai.v32i1.11671},
  url       = {https://doi.org/10.1609/aaai.v32i1.11671}
}

@inproceedings{pei2020iclr-geomgcn,
    title = {{Geom-GCN: Geometric Graph Convolutional Networks}},
    author = {Pei, Hongbin and Wei, Bingzhe and Chang, Kevin Chen-Chuan and Lei, Yu and Yang, Bo},
    booktitle = {International Conference on Learning Representations},
    year = {2020},
    url = {https://mlanthology.org/iclr/2020/pei2020iclr-geomgcn/}
}

@incollection{bridle1990probabilistic,
  author    = {Bridle, John S.},
  title     = {Probabilistic interpretation of feedforward classification network outputs, with relationships to statistical pattern recognition},
  booktitle = {Neurocomputing: Algorithms, Architectures and Applications},
  editor    = {Fogelman Souli{\'e}, Fran{\c{c}}oise and H{\'e}rault, Jeanny},
  series    = {NATO ASI Series},
  volume    = {68},
  pages     = {227--236},
  publisher = {Springer},
  address   = {Berlin, Heidelberg},
  year      = {1990}
}

@inproceedings{Fey/Lenssen/2019,
  title={Fast Graph Representation Learning with {PyTorch Geometric}},
  author={Fey, Matthias and Lenssen, Jan Eric},
  booktitle={ICLR 2019 (RLGM Workshop)},
  year={2019}
}

@inproceedings{clevert2016fast,
  author    = {Djork-Arn{\'e} Clevert and
               Thomas Unterthiner and
               Sepp Hochreiter},
  title     = {Fast and Accurate Deep Network Learning by Exponential Linear Units (ELUs)},
  booktitle = {4th International Conference on Learning Representations, {ICLR} 2016,San Juan, Puerto Rico, May 2-4, 2016, Conference Track Proceedings},
  year      = {2016}
}

@inproceedings{nair2010rectified,
  title={Rectified linear units improve restricted boltzmann machines},
  author={Nair, Vinod and Hinton, Geoffrey E},
  booktitle={Proceedings of the 27th international conference on machine learning (ICML)},
  pages={807--814},
  year={2010}
}

@inproceedings{kingma2015adam,
  author = {Kingma, Diederik P. and Ba, Jimmy},
  title = {Adam: A Method for Stochastic Optimization},
  booktitle = {International Conference on Learning Representations (ICLR)},
  year = {2015},
  url = {https://arxiv.org/abs/1412.6980}
}

@article{hansen1990neural,
  author  = {Hansen, Lars Kai and Salamon, Peter},
  title   = {Neural Network Ensembles},
  journal = {IEEE Transactions on Pattern Analysis and Machine Intelligence},
  volume  = {12},
  number  = {10},
  pages   = {993--1001},
  year    = {1990},
  doi     = {10.1109/34.58871}
}

@article{breiman2001random,
  author  = {Breiman, Leo},
  title   = {Random Forests},
  journal = {Machine Learning},
  year    = {2001},
  volume  = {45},
  number  = {1},
  pages   = {5--32},
  doi     = {10.1023/A:1010933404324}
}

@inproceedings{vaswani2017attention,
  author    = {Vaswani, Ashish and Shazeer, Noam and Parmar, Niki and
               Uszkoreit, Jakob and Jones, Llion and Gomez, Aidan N. and
               Kaiser, {\L}ukasz and Polosukhin, Illia},
  title     = {Attention Is All You Need},
  booktitle = {Advances in Neural Information Processing Systems},
  volume    = {30},
  pages     = {5998--6008},
  year      = {2017},
  publisher = {Curran Associates, Inc.},
  url       = {https://proceedings.neurips.cc/paper/2017/file/3f5ee243547dee91fbd053c1c4a845aa-Paper.pdf}
}

@inproceedings{hu2020open,
  title     = {Open Graph Benchmark: Datasets for Machine Learning on Graphs},
  author    = {Hu, Weihua and Fey, Matthias and Zitnik, Marinka and Dong, Yuxiao and Ren, Hongyu and Liu, Bowen and Catasta, Michele and Leskovec, Jure},
  booktitle = {Advances in Neural Information Processing Systems},
  volume    = {33},
  pages     = {22118--22133},
  year      = {2020}
}

%% else use the following coding to input the bibitems directly in the
%% TeX file.

%% Refer following link for more details about bibliography and citations.
%% https://en.wikibooks.org/wiki/LaTeX/Bibliography_Management

%%\begin{thebibliography}{00}

%% For authoryear reference style
%% \bibitem[Author(year)]{label}
%% Text of bibliographic item

%%\bibitem[Lamport(1994)]{lamport94}
%%  Leslie Lamport,
%%  \textit{\LaTeX: a document preparation system},
 %% Addison Wesley, Massachusetts,
%%  2nd edition,
%%  1994.

%%\end{thebibliography}
\end{document}